\documentclass[10pt,journal,compsoc]{IEEEtran}
\ifCLASSOPTIONcompsoc
  \usepackage[nocompress]{cite}
\else
  \usepackage{cite}
\fi

\usepackage{url}
\usepackage{graphicx}
\usepackage{amsmath} 
\usepackage{bm}
\usepackage{amsfonts,amssymb} 
\usepackage{booktabs}
\usepackage{bbding}
\usepackage{makecell}
\usepackage{xcolor}
\usepackage{wasysym}
\usepackage{multirow}

\usepackage{subcaption}
\usepackage[most]{tcolorbox}

\usepackage{adjustbox}
\usepackage[edges]{forest}

\usepackage{colortbl}
\usepackage[linesnumbered,ruled,vlined]{algorithm2e}
\SetKwComment{Comment}{$\rhd$}{}
\usepackage{enumitem}

\usepackage[pagebackref=false,breaklinks=true,colorlinks,bookmarks=false,urlcolor=magenta]{hyperref}
\newcommand{\myitem}{\noindent\textbf{\emph{$\bullet$}} }
\newcommand{\tunsr}{\textsc{Tunsr }}

\begin{document}

\title{
Towards Unified Neurosymbolic Reasoning \\on Knowledge Graphs}

\author{Qika Lin,~\IEEEmembership{Member,~IEEE,}
        Fangzhi Xu,
        Hao Lu,
        Kai He,
        Rui Mao,\\
        Jun Liu,~\IEEEmembership{Senior Member,~IEEE,}
        Erik Cambria,~\IEEEmembership{Fellow,~IEEE,}
        Mengling Feng,~\IEEEmembership{Senior Member,~IEEE}
\IEEEcompsocitemizethanks{
\IEEEcompsocthanksitem Qika Lin, Kai He, and Mengling Feng are with the Saw Swee Hock School of Public Health, National University of Singapore, 117549, Singapore.
\IEEEcompsocthanksitem Fangzhi Xu and Jun Liu are with the School of Computer Science and Technology, Xi’an Jiaotong University, Xi’an, Shaanxi 710049, China.
\IEEEcompsocthanksitem Hao Lu is with the State Key Laboratory of Multimodal Artificial Intelligence Systems, Institute of Automation, Chinese Academy of Sciences, Beijing 100190, China.
\IEEEcompsocthanksitem Rui Mao and Erik Cambria are with the College of Computing and Data Science, Nanyang Technological University, 639798, Singapore.
}
}



\IEEEtitleabstractindextext{
\begin{abstract}
Knowledge Graph (KG) reasoning has received significant attention in the fields of artificial intelligence and knowledge engineering, owing to its ability to autonomously deduce new knowledge and consequently enhance the availability and precision of downstream applications.
However, current methods predominantly concentrate on a single form of neural or symbolic reasoning, failing to effectively integrate the inherent strengths of both approaches.
%
Furthermore, the current prevalent methods primarily focus on addressing a single reasoning scenario, presenting limitations in meeting the diverse demands of real-world reasoning tasks.
Unifying the neural and symbolic methods, as well as diverse reasoning scenarios in one model is challenging as there is a natural representation gap between symbolic rules and neural networks, and diverse scenarios exhibit distinct knowledge structures and specific reasoning objectives.
To address these issues, we propose a unified neurosymbolic reasoning framework, namely \textsc{Tunsr}, for KG reasoning.
\textsc{Tunsr} first introduces a consistent structure of reasoning graph that starts from the query entity and constantly expands subsequent nodes by iteratively searching posterior neighbors.
Based on it, a forward logic message-passing mechanism is proposed to update both the propositional representations and attentions, as well as first-order logic (FOL) representations and attentions of each node.
In this way, \textsc{Tunsr} conducts the transformation of merging multiple rules by merging possible relations at each step. Finally, the FARI algorithm is proposed to induce FOL rules by constantly performing attention calculations over the reasoning graph.
Extensive experimental results on 19 datasets of four reasoning scenarios (transductive, inductive, interpolation, and extrapolation) demonstrate the effectiveness of \textsc{Tunsr}.
\end{abstract}

}
\maketitle
\IEEEdisplaynontitleabstractindextext
\IEEEpeerreviewmaketitle

\section{Introduction}

\begin{figure*}[t]
\centering
\includegraphics[width=0.9\linewidth]{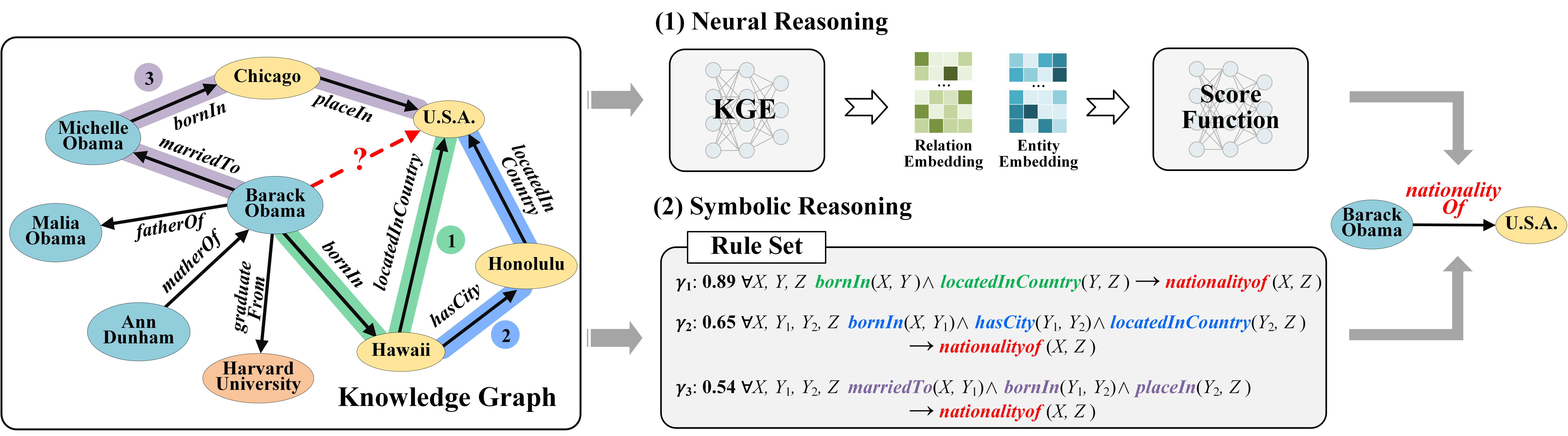}
\caption{Illustration of neural and symbolic methods for KG reasoning. Neural methods learn entity and relation embeddings to calculate the validity of the specific fact. Symbolic methods perform logic deduction using known facts on learned or given rules (like $\gamma_1$, $\gamma_2$ and $\gamma_3$) for inference. }
\label{fig_intro}
\end{figure*}

As a fundamental and significant topic in the domains of knowledge engineering and artificial intelligence (AI),
knowledge graphs (KGs) have been spotlighted in many real-world applications~\cite{tiddi2022knowledge},
such as question answering~\cite{DBLP:conf/aaai/LiM22,DBLP:conf/aaai/MavromatisSIAHG22}, recommendation systems~\cite{DBLP:conf/sigir/YangHXL22,Zhu2023RecommendingLO}, relation extraction~\cite{DBLP:conf/www/BastosN0MSHK21,DBLP:conf/www/ChenZXDYTHSC22} and text generation\cite{DBLP:journals/pami/TrisedyaQWZ22,DBLP:journals/csur/YuZLHWJJ22}.
Thanks to their structured manner of knowledge storage, KGs can effectively capture and represent rich semantic associations between real entities using multi-relational graphical structures.
Factual knowledge is often stored in KGs using the fact triple as the fundamental unit, represented in the form of (\emph{subject}, \emph{relation}, \emph{object}),
such as (\emph{Barack Obama}, \emph{bornIn}, \emph{Hawaii}) in Figure~\ref{fig_intro}.
However, most common KGs, such as Freebase~\cite{DBLP:conf/sigmod/BollackerEPST08} and Wikidata~\cite{DBLP:conf/www/Vrandecic12}, are incomplete due to the limitations of current human resources and technical conditions.
Furthermore, incomplete KGs can degrade the accuracy of downstream intelligent applications or produce completely wrong answers.
Therefore, inferring missing facts from the observed ones is of great significance for downstream KG applications, which is called link prediction that is one form of KG reasoning~\cite{DBLP:journals/tkde/WangMWG17,rossi2021knowledge}.

The task of KG reasoning is to infer or predict new facts using existing knowledge.
For instance, in Figure~\ref{fig_intro}, KG reasoning involves predicting the validity of the target missing triple (\emph{Barack Obama}, \emph{nationalityOf}, \emph{U.S.A.}) based on other available triples.
Using two distinct paradigms, connectionism, and symbolicism, which serve as the foundation for implementing AI systems~\cite{pinker1988connections,trinh2024solving}, existing methods can be categorized into neural, symbolic, and neurosymbolic models.

Neural methods, drawing inspiration from the connectionism of AI, typically employ neural networks
to learn entity and relation representations.
Subsequently, a customized scoring function, such as translation-based distance or semantic matching strategy, is utilized for model optimization and query reasoning, which is illustrated in the top part of Figure~\ref{fig_intro}.
However, such an approach lacks transparency and interpretability~\cite{Lin2023ContrastiveGR,DBLP:conf/acl/XuWSRYYLQ024}.
On the other hand, symbolic methods draw inspiration from the idea of symbolicism in AI.
As shown in the bottom part of Figure~\ref{fig_intro}, they first learn logic rules and then apply these rules, based on known facts to deduce new knowledge.
In this way, symbolic methods offer natural interpretability due to the incorporation of logical rules.
However, owing to the limited modeling capacity given by discrete representation and reasoning strategies of logical rules,
these methods often fall short in terms of reasoning performance~\cite{lin2023fusing}.

\begin{table}[]
\centering
\caption{Classical studies for KG reasoning. \emph{PL} and \emph{FOL} denote the propositional and FOL reasoning, respectively. SKG$_T$, SKG$_I$, TKG$_I$, and TKG$_E$ represent transductive, inductive, interpolation, and extrapolation reasoning. ``$\checkmark$'' means the utilized reasoning manners (neural and logic) or their vanilla application scenarios.}
\resizebox{0.5\textwidth}{!}{
\begin{tabular}{rccccccc}
\toprule
\multirow{2.5}{*}{Model} &\multirow{2.5}{*}{Neural}&\multicolumn{2}{c}{Logic} & \multicolumn{4}{c}{Reasoning Scenarios} \\
\cmidrule(r){3-4}\cmidrule(r){5-8}
&& PL   &FOL   & SKG$_T$          & SKG$_I$    & TKG$_I$ & TKG$_E$ \\
\midrule
TransE~\cite{DBLP:conf/nips/BordesUGWY13}  &\checkmark&   & &\checkmark&&& \\
AMIE~\cite{DBLP:conf/www/GalarragaTHS13}  &&   &\checkmark&\checkmark&&& \\
Neural LP~\cite{DBLP:conf/nips/YangYC17}  &\checkmark&   & \checkmark&\checkmark&&& \\
TAPR~\cite{shen2020modeling}  &\checkmark& \checkmark & &\checkmark&&& \\
RLogic~\cite{DBLP:conf/kdd/ChengL0S22}&\checkmark&&\checkmark&\checkmark&&&\\
\textsc{LatentLogic}~\cite{liu2023latentlogic}&\checkmark&&\checkmark&\checkmark&&&\\
PSRL~\cite{DBLP:conf/eacl/JiangZZLDHL23}&\checkmark&\checkmark&&\checkmark&&&\\
ConGLR~\cite{DBLP:conf/sigir/LinLXPZZZ22}  &\checkmark&   & \checkmark &&\checkmark&& \\
TeAST~\cite{DBLP:conf/acl/LiSG23}&\checkmark&   & &&&\checkmark& \\
TLogic~\cite{DBLP:conf/aaai/LiuMHJT22}  &&   & \checkmark &&&&\checkmark \\
TR-Rules~\cite{DBLP:conf/emnlp/LiELSYSWL23}  &&   & \checkmark &&&& \checkmark\\
TECHS~\cite{DBLP:conf/acl/LinL0XC23}  &\checkmark& \checkmark   & \checkmark&&&&\checkmark \\
\midrule
\cellcolor{gray!50}\textsc{Tunsr}   &\cellcolor{gray!50}\checkmark& \cellcolor{gray!50}\checkmark &\cellcolor{gray!50}\checkmark&\cellcolor{gray!50}\checkmark&\cellcolor{gray!50}\checkmark&\cellcolor{gray!50}\checkmark&\cellcolor{gray!50}\checkmark \\
\bottomrule
\end{tabular}
}
\label{tab_studies}
\end{table}

To leverage the strengths of both neural and symbolic methods while mitigating their respective drawbacks, there has been a growing interest in integrating them to realize neurosymbolic systems~\cite{camnt7}.
Several approaches such as Neural LP~\cite{DBLP:conf/nips/YangYC17}, DRUM~\cite{DBLP:conf/nips/SadeghianADW19}, RNNLogic~\cite{DBLP:conf/iclr/QuCXBT21}, and RLogic~\cite{DBLP:conf/kdd/ChengL0S22} have emerged to address the learning and reasoning of rules by incorporating neural networks into the whole process.
Despite achieving some successes, there remains a notable absence of a cohesive modeling approach that integrates both propositional and first-order logic (FOL) reasoning.
Propositional reasoning on KGs, generally known as multi-hop reasoning~\cite{DBLP:conf/emnlp/ZhangLJLWJY21}, is dependent on entities and predicts answers through specific reasoning paths, which demonstrates strong modeling capabilities by providing diverse reasoning patterns for complex scenarios~\cite{DBLP:conf/acl/ZhangZY000C22,DBLP:journals/tkde/LanHJJZW23}.
On the other hand, FOL reasoning utilizes learned FOL rules to infer information from the entire KG by variable grounding, ultimately scoring candidates by aggregating all possible FOL rules.
FOL reasoning is entity-independent and exhibits good transferability.
Unfortunately, as shown in Table~\ref{tab_studies}, mainstream methods have failed to effectively combine these two reasoning approaches within a single framework, resulting in suboptimal models.

Moreover, as time progresses and society undergoes continuous development, a wealth of new knowledge consistently emerges.
Consequently, simple reasoning on static KGs (SKGs), i.e., transductive reasoning, can no longer meet the needs of practical applications.
Recently, there has been a gradual shift in the research community's focus toward inductive reasoning with emerging entities on SKGs, as well as interpolation and extrapolation reasoning on temporal KGs (TKGs)~\cite{dong2024temporal} that introduce time information to facts.
%
The latest research, which predominantly concentrated on individual scenarios, proved insufficient in providing a comprehensive approach to address various reasoning scenarios simultaneously. This limitation significantly hampers the model's generalization ability and its practical applicability. To sum up, by comparing the state-of-the-art recent studies on KG reasoning in Table~\ref{tab_studies}, it is observed that none of them has a comprehensive unification across various KG reasoning tasks, either in terms of methodology or application perspective.

The challenges in this domain can be categorized into three main aspects:
(1) There is an inherent disparity between the discrete nature of logic rules and the continuous nature of neural networks, which presents a natural representation gap to be bridged.
Thus, implementing differentiable logical rule learning and reasoning is not directly achievable.
(2) It is intractable to solve the transformation and integration problems for propositional and FOL rules, as they have different semantic representation structures and reasoning mechanisms.
(3) Diverse scenarios on SKGs or TKGs exhibit distinct knowledge structures and specific reasoning objectives.
Consequently, a model tailored for one scenario may encounter difficulties when applied to another.
For example, each fact on SKGs is in a triple form while that of TKGs is quadruple.
Conventional embedding methods for transductive reasoning fail to address inductive reasoning as they do not learn embeddings of emerging entities in the training phase.
Similarly, methods employed for interpolation reasoning cannot be directly applied to extrapolation reasoning, as extrapolation involves predicting facts with future timestamps that are not present in the training set.

To address the above challenges, we propose a unified neurosymbolic reasoning framework (named \textsc{Tunsr}) for KG reasoning.
Firstly, to realize the unified reasoning on different scenarios, we introduce a consistent structure of reasoning graph.
It starts from the query entity and constantly expands subsequent nodes (entities for SKGs and entity-time pairs for TKGs) by iteratively searching posterior neighbors.
Upon this, we can seamlessly integrate diverse reasoning scenarios within a unified computational framework, while also implementing different types of propositional and FOL rule-based reasoning over it.
Secondly, to combine neural and symbolic reasoning, we propose a forward logic message-passing mechanism.
For each node in the reasoning graph, \tunsr learns an entity-dependent propositional representation and attention using the preceding counterparts.
Besides, it utilizes a gated recurrent unit (GRU)~\cite{DBLP:journals/corr/ChungGCB14} to integrate the current relation and preceding FOL representations as the edges' representations, following which the entity-independent FOL representation and attention are calculated by message aggregation.
In this process, the information and confidence of the preceding nodes in the reasoning graph are passed to the subsequent nodes and realize the unified neurosymbolic calculation.
Finally, with the reasoning graph and learned attention weights, a novel Forward Attentive Rule Induction (FARI) algorithm is proposed to induce different types of FOL rules.
FARI gradually appends rule bodies by searching over the reasoning graph and viewing the FOL attentions as rule confidences.
It is noted that our reasoning form for link prediction is data-driven to learn rules and utilizes grounding to calculate the fact probabilities, while classic Datalog~\cite{abiteboul1995foundations} and ASP (Answer Set Programming) reasoners~\cite{gebser2011potassco,alviano2013wasp} usually employ declarative logic programming to conduct precise and deterministic deductive reasoning on a set of rules and facts.

In summary, the contribution can be summarized as threefold:

\myitem Combining the advantages of connectionism and symbolicism of AI, we propose a unified neurosymbolic framework for KG reasoning from both perspectives of methodology and reasoning scenarios. To the best of our knowledge, this is the first attempt to do such a study.

\myitem A forward logic message-passing mechanism is proposed to update both the propositional representations and attentions, as well as FOL representations and attentions of each node in the expanding reasoning graph. Meanwhile, a novel FARI algorithm is introduced to induce FOL rules using learned attentions.

\myitem Extensive experiments are carried out on the current mainstream KG reasoning scenarios, including transductive, inductive, interpolation, and extrapolation reasoning. The results demonstrate the effectiveness of our \textsc{Tunsr} and verify its interpretability.

This study is an extension of our model TECHS~\cite{DBLP:conf/acl/LinL0XC23} published at the ACL 2023 conference.
Compared with it, \tunsr has been enhanced in three significant ways:
(1) From the theoretical perspective, although propositional and FOL reasoning are integrated in TECHS for extrapolation reasoning on TKGs,
these two reasoning types are entangled together in the forward process, which limits the interpretability of the model.
However, the newly proposed \textsc{Tunsr} framework presents a distinct separation of propositional and FOL reasoning in each reasoning step.
Finally, they are combined for the reasoning results.
This transformation enhances the interpretability of the model from both propositional and FOL rules' perspectives.
(2) For the perspective of FOL rule modeling, not limited to modeling temporal extrapolation Horn rules in TECHS, the connected and closed Horn rules, and the temporal interpolation Horn rules are also included in the \textsc{Tunsr} framework.
(3) From the application perspective, the TECHS model is customized for the extrapolation reasoning on TKGs.
Based on the further formalization of the reasoning graph and FOL rules, we can utilize the \textsc{Tunsr} model for current mainstream reasoning scenarios of KGs,
including transductive, inductive, interpolation, and extrapolation reasoning.
The experimental results demonstrate that our \textsc{Tunsr} model performs well in all those scenarios.

\section{Preliminaries}

\subsection{KGs, Variants, and Reasoning Scenarios}

Generally, a static KG (SKG) can be represented as $\mathcal{G} = \{\mathcal{E},\mathcal{R},\mathcal{F}\}$, where $\mathcal{E}$ and $\mathcal{R}$ denote the set of entities and relations, respectively. $\mathcal{F}\subset \mathcal{E}\times\mathcal{R}\times\mathcal{E}$ is the fact set. Each fact is a triple, such as ($s$, $r$, $o$), where $s$, $r$, and $o$ denote the head entity, relation, and tail entity, respectively.
By introducing time information in the knowledge, a TKG can be represented as $\mathcal{G}=\{\mathcal{E},\mathcal{R},\mathcal{T},\mathcal{F}\}$, where $\mathcal{T}$ denotes the set of time representations (timestamps or time intervals).
$\mathcal{F}\subset\mathcal{E}\times\mathcal{R}\times\mathcal{E}\times\mathcal{T}$ is the fact set.
Each fact is a quadruple, such as $(s,r,o,t)$ where $s, o\in\mathcal{E}$, $r\in\mathcal{R}$, and $t\in\mathcal{T}$.

\begin{figure}[t]
  \centering
  \begin{minipage}[t]{0.49\linewidth}
    \large
    \centering
    \includegraphics[scale=0.027]{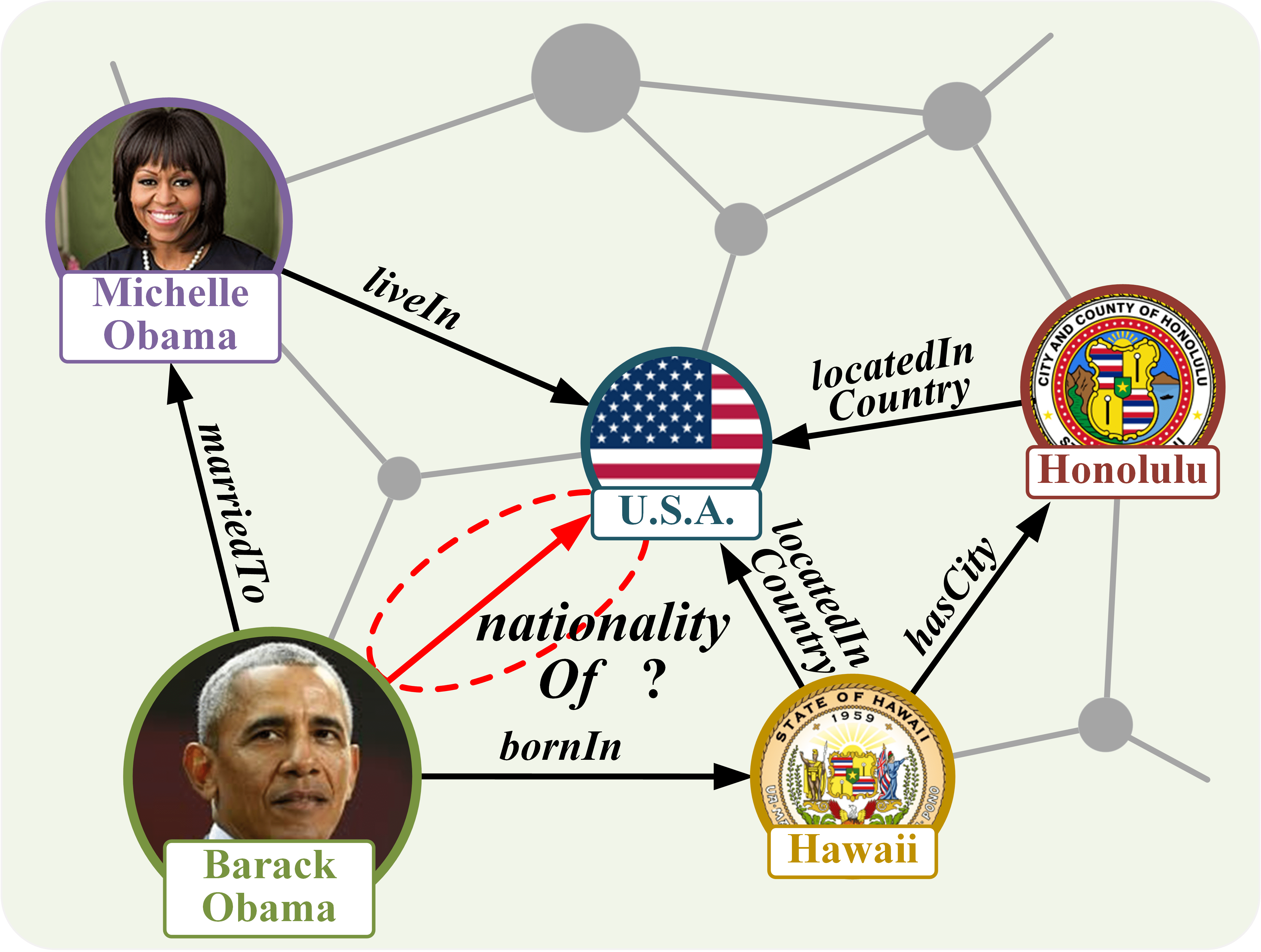}
    \subcaption{Transductive reasoning on SKGs.}
    \label{fig_transductive}
  \end{minipage}
  \begin{minipage}[t]{0.49\linewidth}
    \large
    \centering
    \includegraphics[scale=0.027]{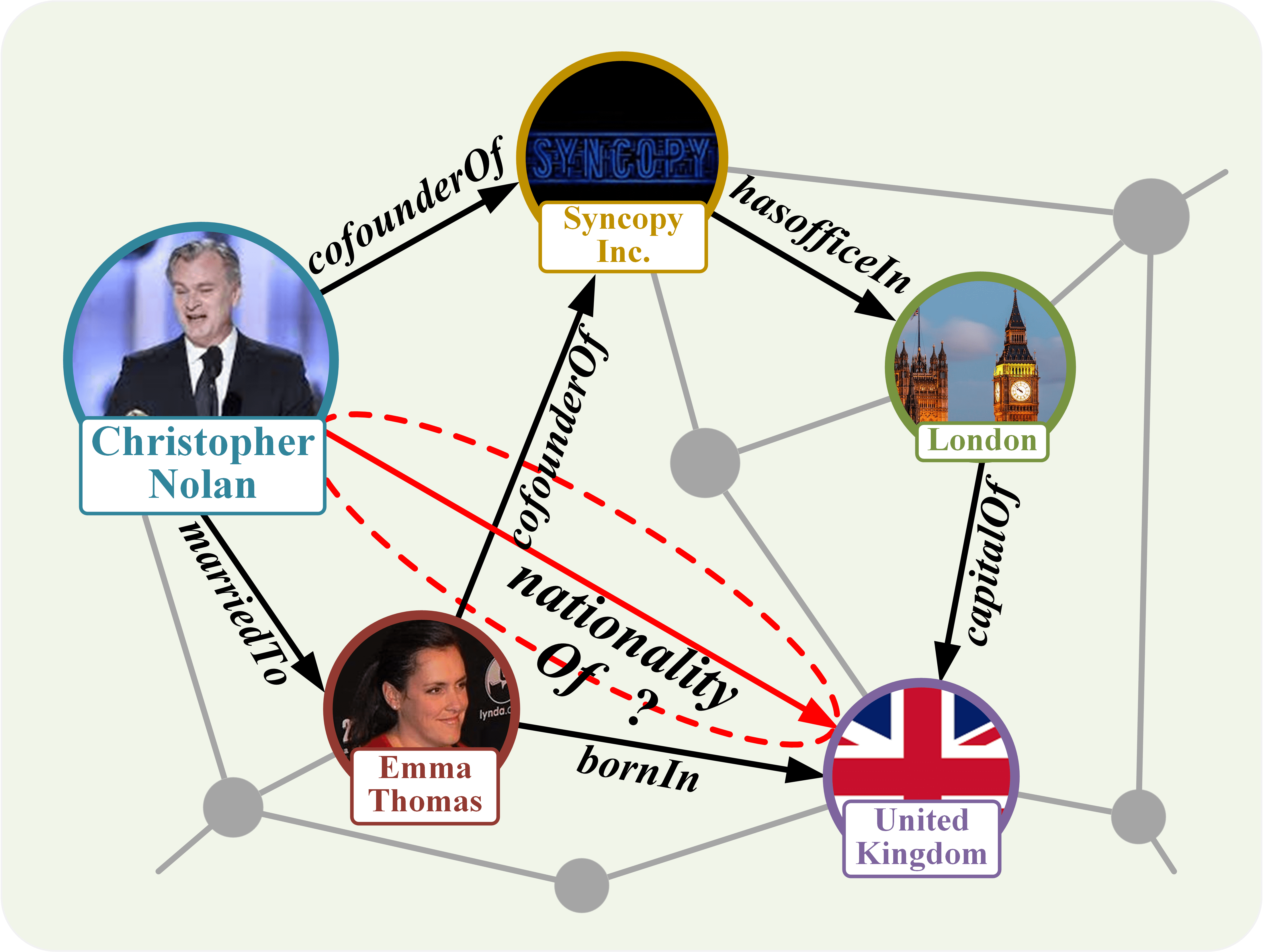}
    \subcaption{Inductive reasoning on SKGs using training data in~\ref{fig_transductive}.}
    \label{fig_inductive}
  \end{minipage}
  \begin{minipage}[t]{0.99\linewidth}
    \large
    \centering
    \includegraphics[scale=0.019]{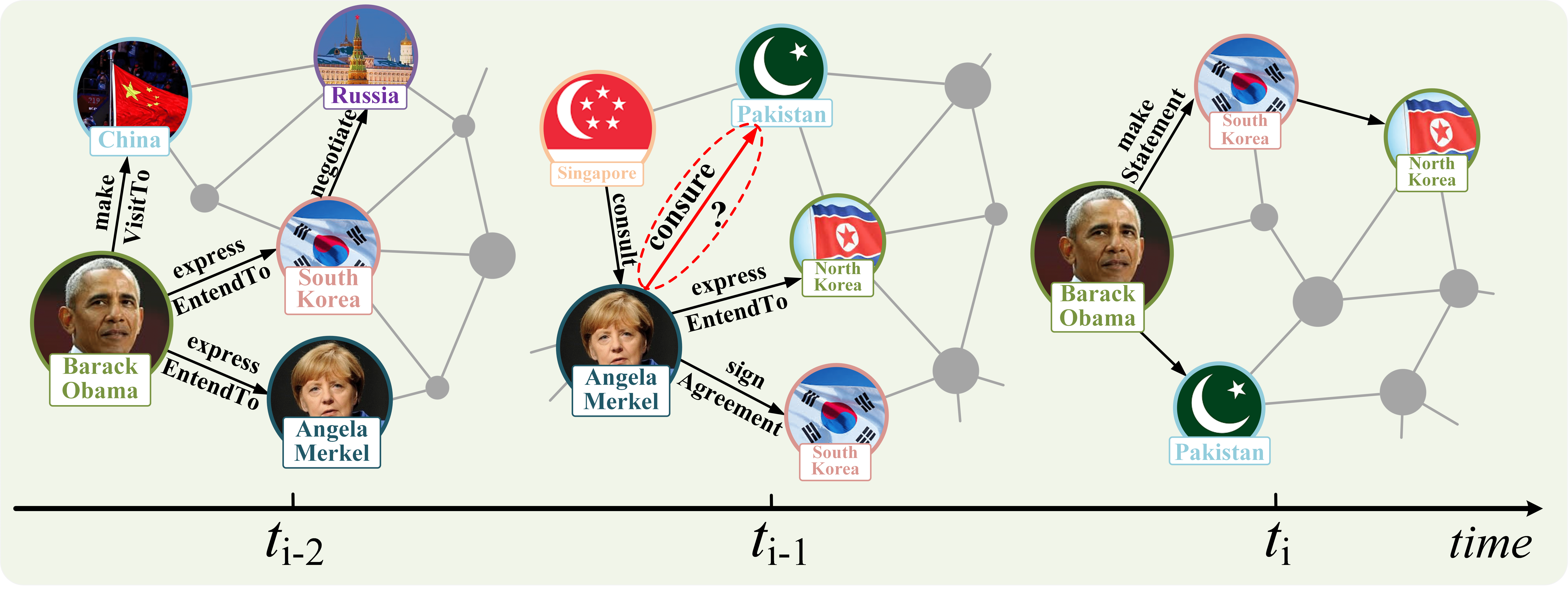}
    \subcaption{Interpolation reasoning on TKGs.}
    \label{fig_interpolation}
  \end{minipage}
  \begin{minipage}[t]{0.99\linewidth}
    \large
    \centering
    \includegraphics[scale=0.019]{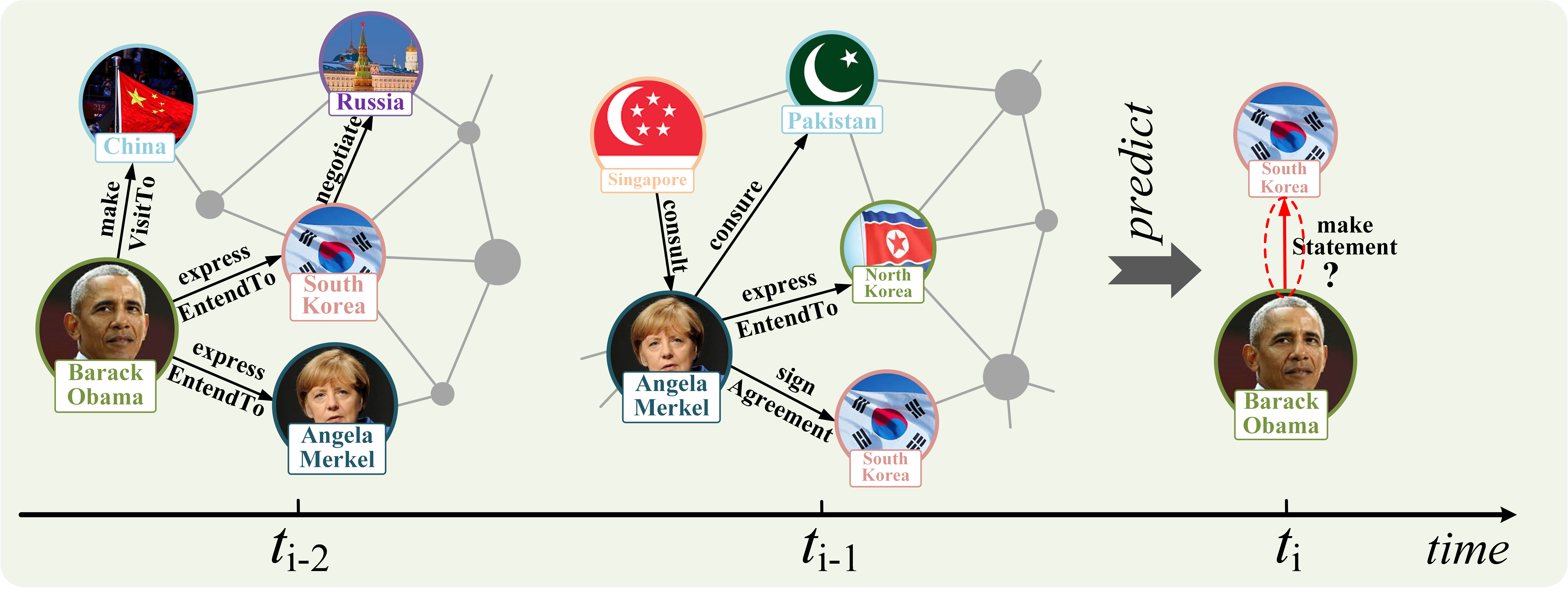}
    \subcaption{Extrapolation reasoning on TKGs.}
    \label{fig_extrapolation}
  \end{minipage}
  \caption{Illustration of four reasoning scenarios on KGs: transductive, inductive, interpolation, and extrapolation. The red dashed arrows indicate the query fact to be predicted.}
  \label{fig_reasoning}
\end{figure}

For these two types of KGs, there are mainly the following reasoning types (query for predicting the head entity can be converted to the tail entity prediction by adding reverse relations), which is illustrated in Figure~\ref{fig_reasoning}:

\myitem Transductive Reasoning on SKGs: Given a background SKG $\mathcal{G} = \{\mathcal{E},\mathcal{R},\mathcal{F}\}$, the task is to predict the missing entity for the query $(\tilde{s},\tilde{r},?)$. The true answer $\tilde{o}\in \mathcal{E}$, and $\tilde{s}\in \mathcal{E}$, $\tilde{r}\in \mathcal{R}$, $(\tilde{s},\tilde{r},\tilde{o}) \notin\mathcal{F}$.

\myitem Inductive Reasoning on SKGs: It indicates that there are new entities appearing in the testing stage, which were not present during the training phase. Formally, the training graph can be expressed as $\mathcal{G}_t = \{\mathcal{E}_t,\mathcal{R},\mathcal{F}_t\}$.
  The inductive graph $\mathcal{G}_i = \{\mathcal{E}_i,\mathcal{R},\mathcal{F}_i\}$ shares the same relation set with $\mathcal{G}_t$. However, their entity sets are disjoint, i.e., $\mathcal{E}_t\cap\mathcal{E}_i=\varnothing$. A model needs to predict the missing entity $\tilde{o}$ for the query $(\tilde{s},\tilde{r},?)$, where $\tilde{s}\in \mathcal{E}_i$, $\tilde{o}\in \mathcal{E}_i$, $\tilde{r}\in \mathcal{R}$, and $(\tilde{s},\tilde{r},\tilde{o}) \notin\mathcal{F}_i$.

\myitem Interpolation Reasoning on TKGs: For a query $(\tilde{s},\tilde{r},?,\tilde{t})$ in the testing phase based on a training TKG $\mathcal{G}_t=\{\mathcal{E}_t,\mathcal{R}_t,\mathcal{T}_t,\mathcal{F}_t\}$,
  a model needs to predict the answer entity $\tilde{o}$ using the facts in the TKG. It denotes that $min(\mathcal{T}_t)\leqslant \tilde{t}\leqslant max(\mathcal{T}_t)$, where $min$ and $max$ denote the functions to obtain the minimum and maximum timestamp within the set, respectively.
  Also, the query satisfies $\tilde{s}\in \mathcal{E}_t$, $\tilde{o}\in \mathcal{E}_t$, $\tilde{r}\in \mathcal{R}_t$, and $(\tilde{s},\tilde{r},\tilde{o},\tilde{t}) \notin\mathcal{F}_t$.

\myitem Extrapolation Reasoning on TKGs: It is similar to the interpolation reasoning that predicts the target entity $\tilde{o}$ for a query $(\tilde{s},\tilde{r},?,\tilde{t})$ in the testing phase, based on a training TKG $\mathcal{G}_t=\{\mathcal{E}_t,\mathcal{R}_t,\mathcal{T}_t,\mathcal{F}_t\}$.
  Differently, this task is to predict future facts, which means the prediction utilizes the facts that occur earlier than $\tilde{t}$ in TKGs, i.e., $\tilde{t}> max(\mathcal{T}_{t})$.


\subsection{Logic Reasoning on KGs}

Logical reasoning involves using a given set of facts (i.e., premises) to deduce new facts (i.e., conclusions) by a rigorous form of thinking~\cite{rautenberg2006concise,ciravegna2023logic}.
It generally covers propositional and first-order logic (also known as predicate logic).
Propositional logic deals with declarative sentences that can be definitively assigned a truth value, leaving no room for ambiguity.
It is usually known as multi-hop reasoning~\cite{DBLP:conf/nips/RenL20,DBLP:conf/acl/ZhangZY000C22} on KGs,
which views each fact as a declarative sentence and usually reasons over query-related paths to obtain an answer.
Thus, propositional reasoning on KGs is entity-dependent. 
First-order logic (FOL) can be regarded as an expansion of propositional logic, enabling the expression of more refined and nuanced ideas~\cite{rautenberg2006concise,andrews2013introduction}.
FOL rules extend the modeling scope and application prospect by introducing quantifiers ($\exists$ and $\forall$), predicates, and variables.
They encompass variables that belong to a specific domain and encompass objects and relationships among those objects~\cite{sun2023survey}.
They are usually in the form of $premise \rightarrow conclusion$, where $premise$ and $conclusion$ denote the rule body and rule head which are all composed of atomic formulas.
Each atomic formula consists of a predicate and several variables, e.g., $bornIn(X, Y)$ in $\gamma_1$ of Figure~\ref{fig_intro}, where $bornIn$ is the predicate and $X$ and $Y$ are all entity variables.
Thus, FOL reasoning is entity-independent, leveraging consistent FOL rules for different entities~\cite{DBLP:journals/corr/abs-2202-07412}.
In this paper, we utilize Horn rules~\cite{DBLP:journals/ai/Poole93} to enhance the adaptability of FOL rules to various KG reasoning tasks. These rules entail setting the rule head to a single atomic formula. Furthermore, to make the Horn rules suitable for multiple reasoning scenarios, we introduce the following definitions.

\noindent \textbf{Connected and Closed Horn (CCH) Rule.}
Based on Horn rules, CCH rules possess two distinct features, i.e., \emph{connected} and \emph{closed}.
The term \emph{connected} means the rule body necessitates a transitive and chained connection between atomic formulas through shared variables.
Concurrently, the term \emph{closed} indicates the rule body and rule head utilize identical start and end variables.

CCH rules of length $n$ (the quantifier $\forall$ would be omitted for better exhibition in the following parts of the paper) are in the following form:
\begin{equation}
\label{cch_rule}
\begin{split}
    \epsilon,\; \forall &X,Y_1,Y_2,\cdots,Y_n,Z\;\;r_1(X, Y_1)\land r_2(Y_1, Y_2)\land \cdots\\ 
    &\land r_{n}(Y_{n-1}, Z)\rightarrow r(X, Z),
\end{split}
\end{equation}
where atomic formulas in the rule body are connected by variables ($X, Y_1, Y_2, \cdots, Y_{n-1}, Z$). For example, $r_1(X, Y_1)$ and $r_2(Y_1, Y_2)$ are connected by $Y_1$.
Meanwhile, all variables form a path from $X$ to $Z$ that are the start variable and end variable of rule head $r_t(X, Z)$, respectively.
$r_1, r_2, \cdots, r_{n}, r$ are relations in KGs to represent predicates.
To model different credibility of different rules, we configure a rule confidence $\epsilon\in [0,1]$ for each Horn rule.
Rule length refers to the number of atomic formulas in the rule body.
For example, $\gamma_1$, $\gamma_2$, and $\gamma_3$ in Figure~\ref{fig_intro} are three example Horn rules of lengths 2, 3, and 3.
Rule grounding of a Horn rule can be realized by replacing each variable with a real entity, e.g., \emph{bornIn(Barack Obama, Hawaii)} $\land$ \emph{locatedInCountry(Hawaii, U.S.A.)} $\rightarrow$ \emph{nationalityOf(Barack Obama, U.S.A.)} is a grounding of rule $\gamma_1$.
CCH rules can be utilized for transductive and inductive reasoning.

\noindent \textbf{Temporal Interpolation Horn (TIH) Rule.} Based on CCH rules on static KGs that require \emph{connected} and \emph{closed} variables, 
TIH rules assign each atomic formula a time variable.

An example of TIH rule can be:
\begin{equation}
\label{tih_rule}
\epsilon,\; \forall X,Y,Z\;\;r_1(X, Y):t_1\land r_2(Y,Z):t_2\rightarrow r(X,Z):t,
\end{equation}
where $t_1$, $t_2$ and $t$ are time variables.
To expand the model capacity when grounding TIH rules, time variables are virtual and do not have to be instantiated to real timestamps, which is distinct from the entity variables (e.g., $X$, $Y$, $Z$).
However, we model the relative sequence of occurrence. This implies that TIH rules with the same atomic formulas but varying time variable conditions are distinct and may have different degrees of confidence, such as for $t_1 <t_2$ vs. $t_1 >t_2$.

\noindent \textbf{Temporal Extrapolation Horn (TEH) Rule.} Based on CCH rules on static KGs that require \emph{connected} and \emph{closed} variables, TEH rules assign each atomic formula a time variable.
Unlike TIH rules, TEH rules have the characteristic of \emph{time growth}, which means the time sequence is increasing and the time in the rule head is the maximum.

For example, the following rule is a TEH rule with length 2:
\begin{equation}
\label{teh_rule}
\begin{split}
    \epsilon,\; \forall X,Y,Z\;\;&r_1(X, Y):t_1\land r_2(Y,Z):t_2\\
    &\rightarrow r(X,Z):t, \;\; s.t.\;\;t_1 \leqslant t_2<t.
\end{split}
\end{equation}
Noticeably, for rule learning and reasoning, $t_1$, $t_2$ and $t$ are also virtual time variables that are only used to satisfy the time growth and do not have to be instantiated.

\begin{figure*}[t]
\centering
\includegraphics[width=0.99\linewidth]{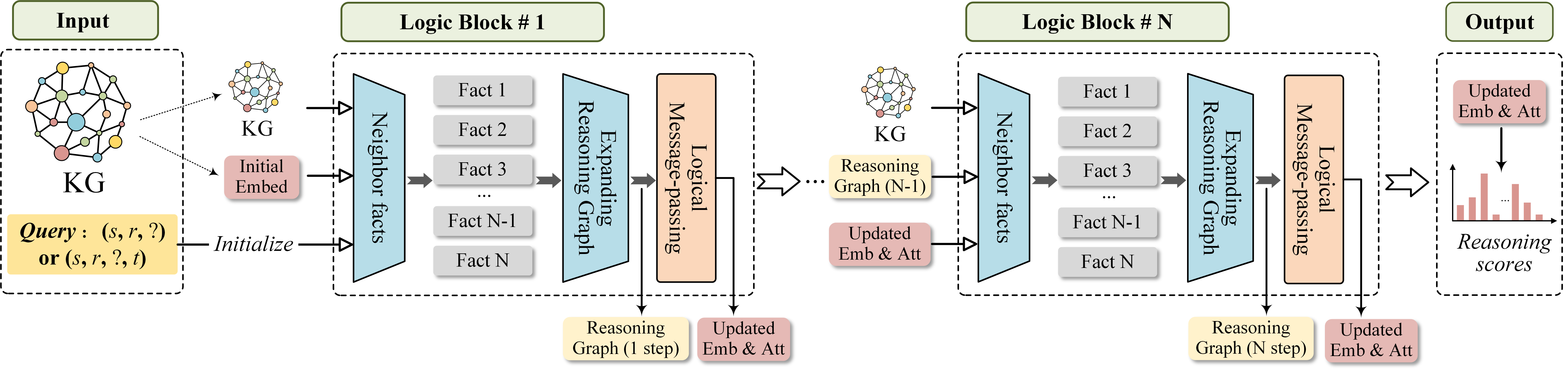}
\caption{An overview of the \textsc{Tunsr}. It utilizes multiple logic blocks to find the answer, where the reasoning graph is constructed and iteratively expanded. Meanwhile, a forward logic message-passing mechanism is proposed to update embeddings and attentions for unified propositional and FOL reasoning.}
\label{fig_arc}
\end{figure*}

\begin{figure}[t]
  \centering
  \begin{minipage}[t]{0.99\linewidth}
    \centering
    \includegraphics[scale=0.59]{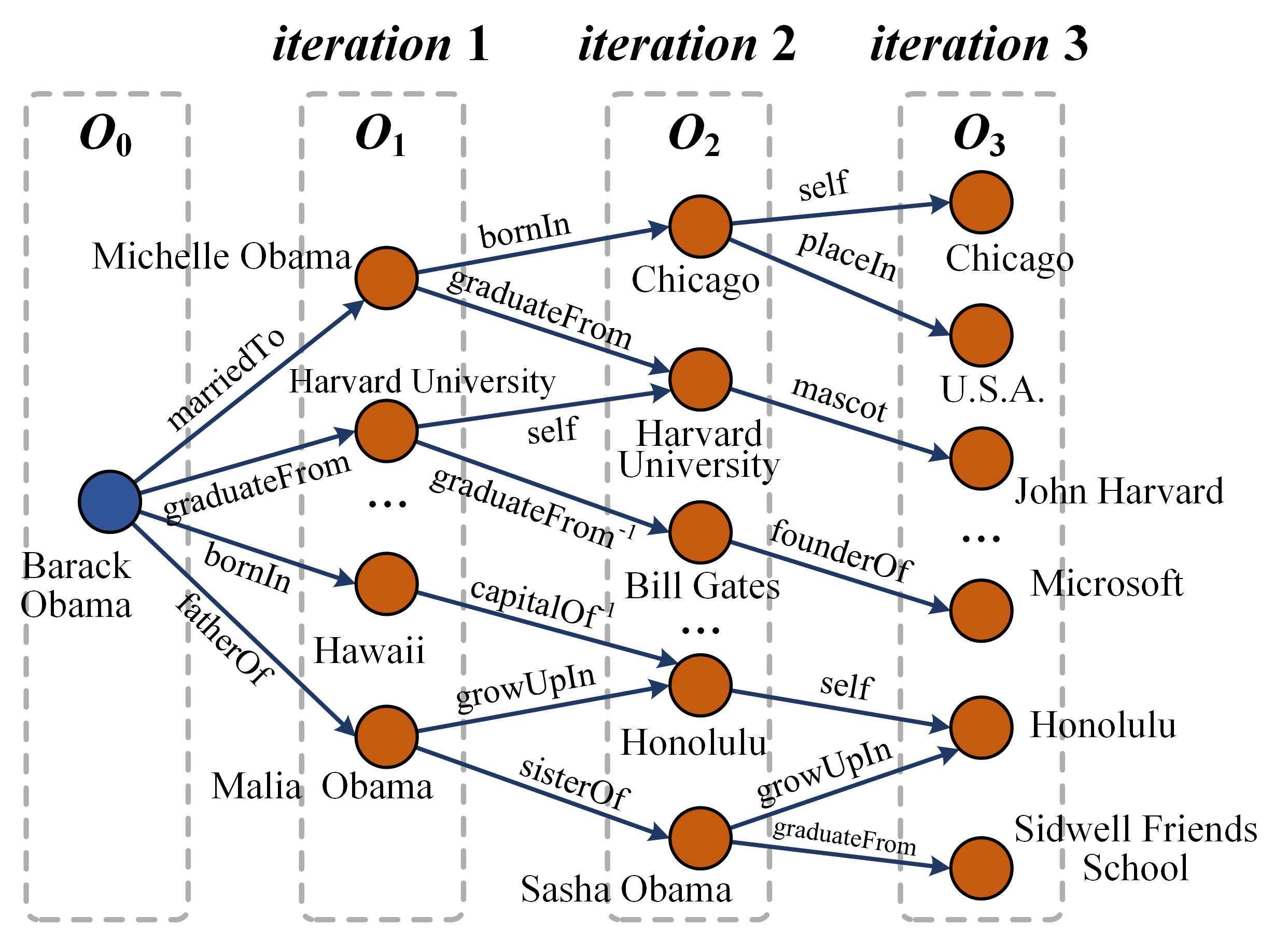}
    \subcaption{An example of reasoning graph in SKGs.}
    \label{fig_rg1}
  \end{minipage}
  \hspace{-1cm}
  \begin{minipage}[t]{0.99\linewidth}
    \centering
    \includegraphics[scale=0.59]{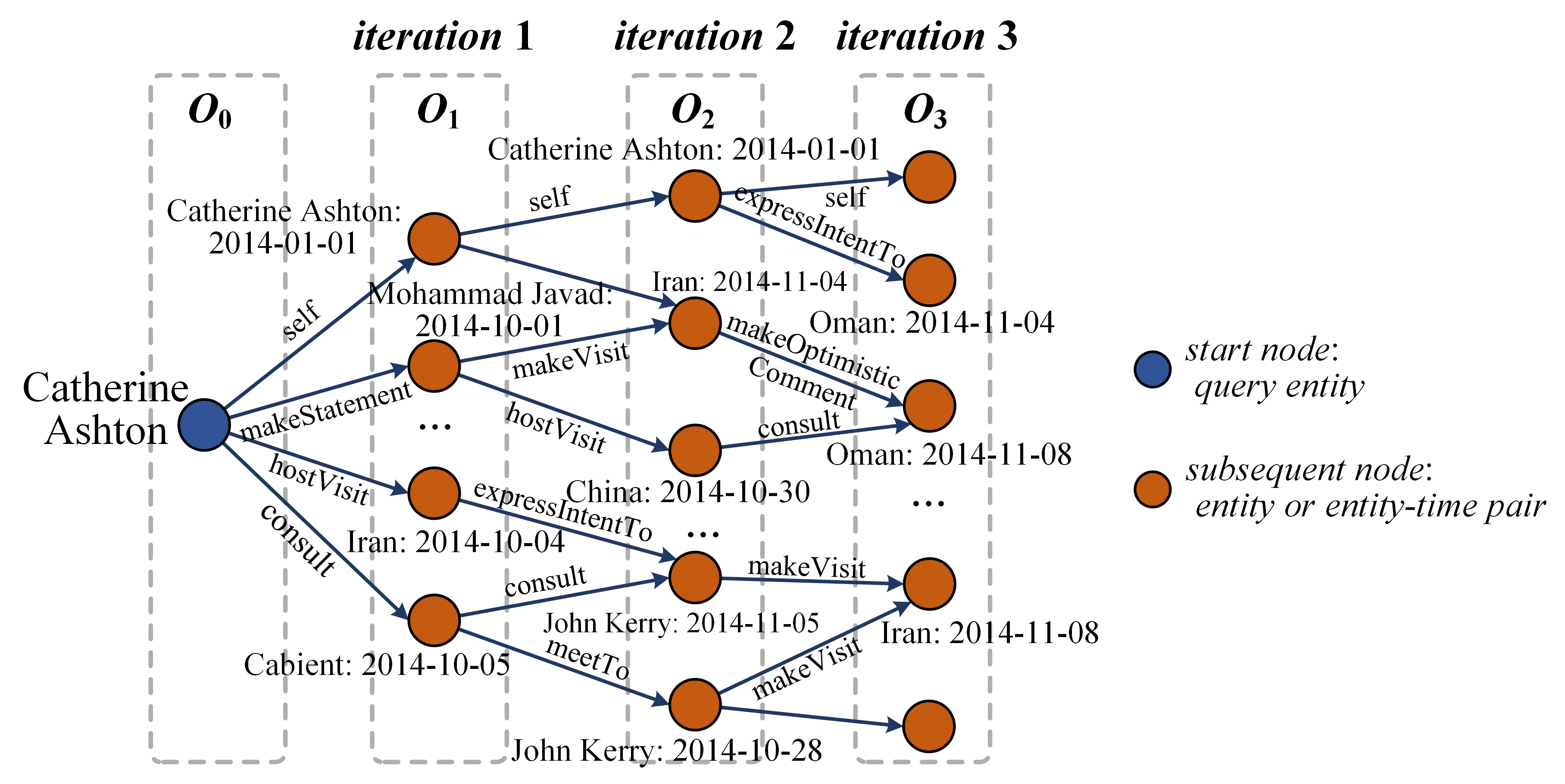}
    \subcaption{An example of reasoning graph in TKGs.}
    \label{fig_rg2}
  \end{minipage}
  \caption{Examples of the reasoning graph with three iterations. (a) is on SKGs while (b) is on TKGs.}
  \label{fig_rg}
\end{figure}

\section{Methodology}
In this section, we present the technical details of our \textsc{Tunsr} model. It leverages a combination of logic blocks to obtain reasoning results, which involves constructing or expanding a reasoning graph and introducing a forward logic message-passing mechanism for propositional and FOL reasoning.
The overall architecture is illustrated in Figure~\ref{fig_arc}.

\subsection{Reasoning Graph Construction}
\label{section_rg}

For each query of KGs, i.e., $\mathcal{Q}=(\tilde{s},\tilde{r},?)$ for SKGs or $\mathcal{Q}=(\tilde{s},\tilde{r},?,\tilde{t})$ for TKGs,
we introduce an expanding reasoning graph to find the answer.
The formulation is as follows.

\noindent \textbf{Reasoning Graph.} For a specific query $\mathcal{Q}$, a reasoning graph is defined as $\widetilde{\mathcal{G}}=\{\mathcal{O}, \mathcal{R}, \widetilde{\mathcal{F}}\}$ for propositional and first-order reasoning.
$\mathcal{O}$ is a node set that consists of nodes in different iteration steps, i.e., $\mathcal{O}=\mathcal{O}_0\cup \mathcal{O}_1\cup\cdots\cup \mathcal{O}_L $.
For SKGs,
$\mathcal{O}_0$ only contains a query entity $\tilde{s}$ and the subsequent is in the form of entities.
$(n_i^l, \bar{r}, n_j^{l+1})\in \widetilde{\mathcal{F}}$ is an edge that links nodes at two neighbor steps, i.e., $n_i^l \in \mathcal{O}_l$, $n_j^{l+1} \in \mathcal{O}_{l+1}$ and $\bar{r} \in \mathcal{R}$.
The reasoning graph is constantly expanded by searching for posterior neighbor nodes.
For start node $n^0=\tilde{s}$, its posterior neighbors are $\mathcal{N}(n^0)=\{e_i|(\tilde{s},\bar{r},e_i)\in \mathcal{F}\}$.
For a node in following steps $n_i^l=e_i\in \mathcal{O}_l$, its posterior neighbors are $\mathcal{N}(n_i^l)=\{e_j|(e_i,\bar{r},e_j)\in \mathcal{F} \}$.
Its preceding parents are $\widetilde{\mathcal{N}}(n_i^l)=\{(n_j^{l-1},\bar{r})|n_j^{l-1}\in \mathcal{O}_{l-1}\land (n_j^{l-1},\bar{r}, n_i^{l})\in \widetilde{\mathcal{F}}\}$.
To take preceding nodes into account at the current step, an extra relation \emph{self} is added.
Then, $n_i^l=e_i$ can be obtained at the next step as $n_i^{l+1}=e_i$ and there have $(n_i^{l}, self, n_i^{l+1})\in \widetilde{\mathcal{F}}$.

For TKGs,
$\mathcal{O}_0$ also contains a query entity $\tilde{s}$.
But the following nodes are in the form of entity-time pairs.
In the interpolation scenarios, for start node $n^0=\tilde{s}$, its posterior neighbors are $\mathcal{N}(n^0)=\{(e_i,t_i)|(\tilde{s},\bar{r},e_i,t_i)\in \mathcal{F} \}$.
For a node in following steps $n_i^l=(e_i,t_i)\in \mathcal{O}_l$, its posterior neighbors are $\mathcal{N}(n_i^l)=\{(e_j,t_j)|(e_i,\bar{r},e_j,t_j)\in \mathcal{F} \}$.
Differently, in the extrapolation scenarios, for start node $n^0=\tilde{s}$, its posterior neighbors are $\mathcal{N}(n^0)=\{(e_i,t_i)|(\tilde{s},\bar{r},e_i,t_i)\in \mathcal{F} \land t_i < \tilde{t}\}$.
For a node in following steps $n_i^l=(e_i,t_i)\in \mathcal{O}_l$, its posterior neighbors are $\mathcal{N}(n_i^l)=\{(e_j,t_j)|(e_i,\bar{r},e_j,t_j)\in \mathcal{F} \land t_i \leqslant t_j \land t_j < \tilde{t}\}$.
Similar to the situation of SKGs,
the preceding parents of nodes in TKG scenarios are also $\widetilde{\mathcal{N}}(n_i^l)=\{(n_j^{l-1},\bar{r})|n_j^{l-1}\in \mathcal{O}_{l-1}\land (n_j^{l-1},\bar{r}, n_i^{l})\in \widetilde{\mathcal{F}}\}$ and an extra relation \emph{self} is also added.
Then, $n_i^l=(e_i,t_i)$ can be obtained at the next step as $n_i^{l+1}=(e_i,t_i)$ ($t_i$ is the minimum time if $l=0$) and there have $(n_i^{l}, self, n_i^{l+1})\in \widetilde{\mathcal{F}}$.

Two examples of the reasoning graph with three iterations are shown in Figure~\ref{fig_rg}.
Through the above processing,
we can model both propositional and FOL reasoning in a unified manner for different reasoning scenarios. 

\subsection{Modeling of Propositional Reasoning}

For decoding the answer for a specific query $\mathcal{Q}$, we introduce an iterative forward message-passing mechanism in a continuously expanding reasoning graph, regulated by propositional and FOL reasoning.
In the reasoning graph, we set two learnable parameters for each node $n_i^l$ to guide the propositional computation:
propositional embedding ${\rm \textbf x}_i^l$ and propositional attention ${\alpha}_{n_i^l}$.
For a better presentation, we employ the reasoning process on TKGs to illustrate our method. SKGs can be considered a specific case of TKGs' when the time information of the nodes in the reasoning graph is removed.
The initialized embeddings for entity, relation, and time are formalized as ${\rm \textbf h}$, ${\rm \textbf g}$, and ${\rm \textbf e}$.
Time embeddings are obtained by the generic time encoding~\cite{DBLP:conf/iclr/XuRKKA20} as it is fully compatible with attention to capture temporal dynamics, which is defined as:
${\rm \textbf e}_t\! =\! \sqrt{\frac{1}{d_t}}[{\rm cos}(w_1t+b_1), \cdots, {\rm cos}(w_{d_t}t+b_{d_t})]$,
where $[w_1, \cdots, w_{d_t}]$ and $[b_1, \cdots, b_{d_t}]$ are trainable parameters for transformation weights and biases.
{\rm cos} denotes the standard cosine function and $d_t$ is the dimension of time embedding.

Further, the start node $n^0$=$\tilde{s}$ is initialized as its embedding ${\rm \textbf x}_{\tilde{s}}={\rm \textbf h}_{\tilde{s}}$.
The node $n_i=(e_i,t_i)$ at the following iterations is firstly represented by the linear transformation of embeddings: ${\rm \textbf x}_i$=${\rm \textbf W}_n[{\rm \textbf h}_{e_i}\|{\rm \textbf e}_{t_i}]$ (${\rm \textbf W}$ represents linear transformation and $\|$ denotes the embedding concatenation in the paper).
Constant forward computation is required in the reasoning sequence of the target when conducting multi-hop propositional reasoning.
Thus, forward message-passing is proposed to pass information (i.e., representations and attention weights) from the preceding nodes to their posterior neighbor nodes. The computation of each node is contextualized with preceding information that contains both entity-dependent parts, reflecting the continuous accumulation of knowledge and credibility in the reasoning process.
Specifically, to update node embeddings in step $l$+1, its own feature and the information from its priors are integrated:
\begin{equation}
\label{eq4}
  {\rm \textbf x}_j^{l+1}={\rm \textbf W}_{1}^{l}{\rm \textbf x}_j+\!\!\!\!\sum_{(n_i^l,\bar{r})\in \widetilde{\mathcal{N}}(n_j^{l+1})}\!\!\!\!\alpha_{n_i^l,\bar{r},n_j^{l+1}}{\rm \textbf W}_{2}^{l}{\rm \textbf m}_{n_i^l,\bar{r},n_j^{l+1}},
\end{equation}
where ${\rm \textbf m}_{n_i^l,\bar{r},n_j^{l+1}}$ is the message from a preceding node to its posterior node, which is given by the node and relation representations:
\begin{equation}
\label{eq5}
  {\rm \textbf m}_{n_i^l,\bar{r},n_j^{l+1}}\!=\!{\rm \textbf W}_{3}^{l}[{\rm \textbf n}_i^l\|{\rm \textbf g}_{\bar{r}}\|{\rm \textbf n}_j].
\end{equation}

This updating form superficially seems similar to the general message-passing in GNNs~\cite{Lin2023ContrastiveGR}.
However, they are actually different as ours is in a one-way and hierarchical manner, which is tailored for the tree-like structure of the reasoning graph.
The propositional attention weight $\alpha_{n_i^l,\bar{r},n_j^{l+1}}$ is for each edge in a reasoning graph.
As propositional reasoning is entity-dependent, we compute it by the semantic association of entity-dependent embeddings between the message and the query:
\begin{equation}
\label{eq6}
  e_{n_i^l,\bar{r},n_j^{l+1}} \!=\! \textsc{sigmoid}( {\rm \textbf W}_{4}^{l}[{\rm \textbf m}_{n_i^l,\bar{r},n_j^{l+1}}\|{\rm \textbf q}]),
\end{equation}
where ${\rm \textbf q}={\rm \textbf W}_{q}[{\rm \textbf h}_{\tilde{s}}\|{\rm \textbf g}_{\tilde{r}}\|{\rm \textbf e}_{\tilde{t}}]$ is the query embedding.
Then, the softmax normalization is utilized to scale edge attentions on this iteration to [0,1]:
\begin{equation}
\label{eq7}
  \alpha_{\!n_i^l,\bar{r},n_j^{l+1}} \!\!=\!\! \frac{\exp(e_{n_i^l,\bar{r},n_j^{l+1}})}{\sum_{(\!n_{i'}^l,\bar{r}')\in \widetilde{\mathcal{N}}(n_j^{l+1}\!)} \!\!\exp(e_{n_{i'}^l,\bar{r}',n_j^{l+1}}\!)},
\end{equation}
Finally, the propositional attention of new node $n_j^{l+1}$ is aggregated from edges for the next iteration:
\begin{equation}
\label{eq8}
\begin{split}
  &\alpha_{n_j^{l+1}} \!=\!\!\!\sum_{(n_i^l,\bar{r})\in \widetilde{\mathcal{N}}(n_j^{l+1})} \!\!\!\!\!\!\!\! \alpha_{n_i^l,\bar{r},n_j^{l+1}}.
\end{split} 
\end{equation}

\subsection{Modeling of FOL Reasoning}

Different from propositional reasoning, FOL reasoning is entity-independent and has a better ability for generalization.
As first-order reasoning focuses on the interaction among entity-independent relations, we first obtain the hidden FOL embedding of an edge by fusing the hidden FOL embedding of the preceding node and current relation representation via a GRU~\cite{DBLP:journals/corr/ChungGCB14}.
Then, the FOL representation ${\rm \textbf y}$ and attention $b$ are given by:
\begin{equation}
\label{eq9}
  {\rm \textbf y}_{n_i^l,\bar{r},n_j^{l+1}}\! =\! \textsc{gru}({\rm \textbf g}_{\bar{r}}, {\rm \textbf y}_{n_i^{l}}),
\end{equation}
\begin{equation}
\label{eq10}
  b_{n_i^l,\bar{r},n_j^{l+1}}\! = \!\textsc{sigmoid} ( {\rm \textbf W}_{5}^{l}{\rm \textbf y}_{n_i^l,\bar{r},n_j^{l+1}}).
\end{equation}
Since the preceding node with high credibility leads to faithful subsequent nodes, the attention of the prior ($\beta$) flows to the current edge.
Then, the softmax normalization is utilized to scale edge attentions on this iteration to [0,1]:
\begin{equation}
\label{eq11}
\begin{split}
  b_{n_i^l,\bar{r},n_j^{l+1}}& = \beta_{\!n_i^l}\cdot b_{n_i^l,\bar{r},n_j^{l+1}}, \;\;\\
  \beta_{\!n_i^l,\bar{r},n_j^{l+1}} \!\!&=\!\! \frac{\exp(b_{n_i^l,\bar{r},n_j^{l+1}})}{\sum_{(\!n_{i'}^l,\bar{r}')\in \widetilde{\mathcal{N}}(n_j^{l+1}\!)} \!\!\exp(b_{n_{i'}^l,\bar{r}',n_j^{l+1}}\!)},
\end{split}
\end{equation}
Finally, the FOL representation and attention of a new node $n_j^{l+1}$ are aggregated from edges for the next iteration:
\begin{equation}
\label{eq12}
\begin{split}
  {\rm \textbf y}_{n_j^{l+1}}\!&=\!\!\! \sum_{(n_i^l,\bar{r})\in \widetilde{\mathcal{N}}(n_j^{l+1})}\!\!\!\!\beta_{n_i^l,\bar{r},n_j^{l+1}} {\rm \textbf y}_{n_i^l,\bar{r},n_j^{l+1}},\\
  &\beta_{n_j^{l+1}} \!=\!\!\!\sum_{(n_i^l,\bar{r})\in \widetilde{\mathcal{N}}(n_j^{l+1})} \!\!\!\!\!\!\!\! \beta_{n_i^l,\bar{r},n_j^{l+1}}.
\end{split} 
\end{equation}

\noindent \textbf{Insights of FOL Rule Learning and Reasoning.}

Actually, \tunsr introduces a novel FOL learning and reasoning strategy by forward logic message-passing mechanism over reasoning graphs.
In general, the learning and reasoning of FOL rules on KGs or TKGs are usually in two-step fashion~\cite{DBLP:conf/www/GalarragaTHS13,DBLP:journals/vldb/GalarragaTHS15,DBLP:conf/www/ZhangPWCZZBC19,DBLP:conf/iclr/QuCXBT21,DBLP:conf/aaai/LiuMHJT22,DBLP:conf/kdd/ChengL0S22,lin2023fusing}.
First, it searches over whole data to mine rules and their confidences.
Second, for a query, the model instantiates all variables to find all groundings of learned rules and then aggregates all confidences of eligible rules.
For example, for a target entity $o$, its score can be the sum of learned rules with valid groundings
and rule confidences can be modeled by a GRU.
However, this is apparently not differentiable and cannot be optimized in an end-to-end manner because of the discrete rule learning and grounding operations. 
Thus, our model conducts the transformation of merging multiple rules by merging possible relations at each step, using FOL attention as:
\begin{equation}
\label{eq13}
\begin{split}
  &\underbrace{S_o\!=\! \sum_{\gamma\in\Gamma} \beta_{\gamma}
  \!=\!\sum_{\gamma\in\Gamma} f\big[\textsc{gru}({\rm \textbf g}_{\gamma,h},{\rm \textbf g}_{\gamma, b^1},\cdots, {\rm \textbf g}_{\gamma, b^{|\gamma|}})]}_{(a)}\\
  &\underbrace{\approx \prod_{l=1}^{L}\sum_{n_j\in\mathcal{O}_l}\bar{f_l}\big[\textsc{gru}({\rm \textbf g}_{\bar{r}}, {\rm \textbf o}_{n_j}^{l})) \big ]}_{(b)}.
\end{split}
\end{equation}
$\beta_\gamma$ is the confidence of rule $\gamma$.
${\rm \textbf g}_{\gamma,h}$ and ${\rm \textbf g}_{\gamma,b^i}$ are the relation embeddings of head $h$ and $i$-th body $b^i$ of this rule.
Part (a) utilizes the grounding of the learned rules to calculate reasoning scores, where each rule's confidence can be modeled by GRU and feedforward network $f$.
We can conduct reasoning at each step rather than whole multi-step processing, so the previous can approximate to part (b).
$\bar{f_l}$ is for the attention calculation.
In this way, the differentiable process is achieved. This is an extension and progression of Neural LP~\cite{DBLP:conf/nips/YangYC17} and DURM~\cite{DBLP:conf/nips/SadeghianADW19} by introducing several specific strategies for unified KG reasoning.
Finally, 
the real FOL rules can be easily induced to constantly perform attention calculation over the reasoning graph, which is summarized as the Forward Attentive Rule Induction (FARI) algorithm. 
It is shown in Algorithm~\ref{algorithm_rule}, where the situation on TKGs is given and that on SKGs can be obtained by omitting time information.
In this way, \textsc{Tunsr} has the ability to capture CCH, TIH, and TEH rules with the specific-designed reasoning graphs as described in Section~\ref{section_rg}.
As we add an extra \emph{self} relation in the reasoning graph, the FARI algorithm can obtain all possible rules (no longer than length \emph{L}) by deleting existing atoms with the \emph{self} relation in induced FOL rules.

\begin{algorithm}[t]
  \KwIn{the reasoning graph $\widetilde{\mathcal{G}}$, FOL attentions $\beta$.
  }
  \KwOut{the FOL rule set $\Gamma$.}
  Init $\Gamma=\varnothing$, $B(n_{\tilde{s}}^0)=[0,[]]$, $\mathcal{D}_{0}[n_{\tilde{s}}^0]=[1, B(n_{\tilde{s}}^0)]$\;
  \For{l=1 to L of decoder iterations} 
    {  Initialize node-rule dictionary $\mathcal{D}_{l}$\;
      \For{node $n_j^l$ in $\mathcal{O}_{l}$}
        {  Set rule body list $B(n_j^l)$ = [] \;
          \For{($n_i^{l-1},\bar{r}$) of $\widetilde{\mathcal{N}}$($n_j^l$) in $\mathcal{O}_{l-1}$}
          {  
            Prior $e_{i,l-1}^2$, $B(n_i^{l-1})$ = $\mathcal{D}_{l-1}[n_i^{l-1}]$\;
            \For{weight $\epsilon$, body $\gamma_b$ in $B(n_i^{l-1})$}
            {
            $\epsilon' = e_{i,l-1}^2 \cdot e_{n_i^{l-1},\bar{r},n_j^{l}}^2$ \;
            \small{$\gamma'_b = \gamma_b.add(\bar{r})$, $B(n_j^l).add([\epsilon',\gamma'_b])$} \;
            }
          }
          $e_{j,l}^2=sum\{[\epsilon \in B(n_j^l)]\}$ \;
          Add $n_j^l$: [$e_{j,l}^2$, $B(n_j^l)$] to $\mathcal{D}_{l}$ \;
        }
      Normalize $e_{j,l}^2$ of $n_j^l$ in $\mathcal{O}_{l}$ using softmax\;
    }
  \For{$n_i^L$ in $\mathcal{O}_{L}$}
  {
  $e_{i,L}^2$, $B(n_i^L)$ = $\mathcal{D}_{L}[n_j^L]$ \;
  \For{$\epsilon, \gamma_b$ in $B(n_i^L)$}
    {
    $\Gamma.add([\epsilon, \gamma_b[1](X,Y_1):t_1 \land \cdots\land \gamma_b[L](Y_{L-1},Z):t_L \rightarrow \tilde{r}(X,Z):t])$
    }
  }
  \textbf{Return} rule set $\Gamma$.
 \caption{FARI for FOL rules Induction.}
 \label{algorithm_rule}
\end{algorithm}

\subsection{Reasoning Prediction and Process Overview}

After calculation with $L$ logic blocks, the reasoning score for each entity can be obtained.
For each entity $o$ at the last step of the reasoning graph for SKGs, we can utilize the representation and attention value of the propositional and FOL reasoning for calculating answer validity:
\begin{equation}
\label{eq14}
  {\rm \textbf h}_o = (1-\lambda){\rm \textbf x}_o + \lambda{\rm \textbf y}_o, \gamma_o = (1-\lambda)\alpha_o + \lambda\beta_o,
\end{equation}
where $\lambda$ is a learnable weight for the combination of propositional and FOL reasoning.
$\alpha_o$ and $\beta_o$ are learned attention values for propositional and FOL reasoning, respectively.
We calculate it dynamically using propositional embedding ${\rm \textbf x}_o$, FOL embedding ${\rm \textbf y}_o$, and query embedding ${\rm \textbf q}$.
Based on it, the final score is given by:
\begin{equation}
\label{eq15}
  s(\mathcal{Q},o) = {\rm \textbf W}_{5}{\rm \textbf h}_o+\gamma_o.
\end{equation}
Reasoning scores for those entities that are not in the last step of the reasoning graph are set to 0 as it indicates that there are no available propositional and FOL rules for those entities.
Finally, the model is optimized by the multi-class log-loss~\cite{DBLP:conf/icml/LacroixUO18} like RED-GNN:
\begin{equation}
\label{eq16}
  \mathcal{L}=\sum_{\mathcal{Q}} \Big[-s(\mathcal{Q},o)+\log\big(\sum_{\bar{o}\in \mathcal{E}}\exp(s(\mathcal{Q},\bar{o}))\big)\Big],
\end{equation}
where $s(\mathcal{Q},o)$ denotes the reasoning score of labeled entity $o$ for query $\mathcal{Q}$, while $\bar{o}$ is the arbitrary entity.
For reasoning situations of TKGs, we need firstly aggregate node embedding and attentions with the same entity to get the entity score.
Because the nodes in the reasoning graph of TKGs except the start node are in the form of entity-time pair.

The number of nodes may explode in the reasoning graph as it shows an exponential increase to reach $|\mathcal{N}(n_i)|^L$ by iterations, especially for TKGs.
For computational efficiency, we introduce the strategies of iteration fusion and sampling for interpolation and extrapolation reasoning, respectively.
In the interpolation scenarios, nodes of entity-time pairs with the same entity are fused to an entity node and then are used to expand the reasoning graph.
In the extrapolation scenarios, 
posterior neighbors of each node are sampled with a maximum of \emph{M} nodes in each iteration.
For sampling \emph{M} node in the reasoning graph, we follow a time-aware weighted sampling strategy, considering that recent events may have a greater impact on the forecast target.
Specifically, for a posterior neighbor node with time $t'$, we compute its sampling weight by $\frac{\exp (t'-\tilde{t})}{\sum_{\bar{t}}{\exp (\bar{t}-\tilde{t})}}$ for the query ($\tilde{s}$,$\tilde{r}$,?,$\tilde{t}$), where $\bar{t}$ denotes the time of all possible posterior neighbor nodes for a prior node. 
After computing attention weights for each edge in the same iteration, we select top-\emph{N} among them with larger attention weights and prune others.

\section{Experiments and Results}

\subsection{Experiment Setups}

The baselines cover a wide range of mainstream techniques and strategies for KG reasoning, with detailed descriptions provided in the Appendix.
In the following parts of this section, we will carry out experiments and analyze results to answer the following four research questions.

\myitem \textbf{RQ1.} How does the unified \textsc{Tunsr} perform in KG reasoning compared to state-of-the-art baselines?

\myitem \textbf{RQ2.} How effective are propositional and FOL reasoning, and is it reasonable to integrate them?

\myitem \textbf{RQ3.} What factors affect the reasoning performance of the \textsc{Tunsr} framework?

\myitem \textbf{RQ4.} What is the actual reasoning process of \textsc{Tunsr}?

\subsection{Comparison Results (RQ1)}

\begin{table*}[t!]
\caption{The experiment results of transductive reasoning. The optimal and suboptimal values on each metric are marked in red and blue, respectively. The percent signs (\%) for Hits@k metrics are omitted for better presentation. The following tables have a similar setting.
}
\centering
\begin{tabular}
{rp{0.93cm}<{\centering}p{0.93cm}<{\centering}p{0.93cm}<{\centering}p{0.93cm}<{\centering}p{0.93cm}<{\centering}p{0.93cm}<{\centering}p{0.93cm}<{\centering}p{0.93cm}<{\centering}}
\toprule
\multicolumn{1}{c}{\multirow{2}{*}{\textbf{Model}}}& \multicolumn{4}{c}{\textbf{WN18RR}}   & \multicolumn{4}{c}{\textbf{FB15k237}}  \\
\cmidrule(r){2-5} \cmidrule(r){6-9}
\multicolumn{1}{c}{}& \textbf{MRR}  & \textbf{Hits@1} & \textbf{Hits@3} & \textbf{Hits@10} & \textbf{MRR}  & \textbf{Hits@1} & \textbf{Hits@3} & \textbf{Hits@10} \\
\midrule 
TransE~\cite{DBLP:conf/nips/BordesUGWY13}& 0.481 &43.30& 48.90& 57.00 &0.342 &24.00& 37.80& 52.70\\
DistMult~\cite{DBLP:journals/corr/YangYHGD14a}& 0.430 &39.00 &44.00 &49.00& 0.241& 15.50& 26.30 &41.90 \\
UltraE~\cite{DBLP:conf/kdd/XiongZNXP0S22} &0.485& 44.20& 50.00& 57.30& 0.349& 25.10& 38.50& 54.10\\
ComplEx-DURA~\cite{DBLP:journals/pami/WangZSCJW23} &0.491 & 44.90 &--&57.10 &0.371& 27.60&--& \textbf{\textcolor{blue}{56.00}}\\
AutoBLM~\cite{DBLP:journals/pami/ZhangYK23} & 0.490 &45.10 &--&56.70 &0.360 &26.70 &--&55.20 \\
SE-GNN~\cite{DBLP:conf/aaai/LiCZB0L022} & 0.484 &44.60 &50.90&57.20 &0.365 &27.10 &39.90&54.90 \\
RED-GNN~\cite{DBLP:conf/www/ZhangY22} & 0.533 &48.50 &--&62.40 &\textbf{\textcolor{blue}{0.374}}&28.30 &--&55.80\\
CompoundE~\cite{ge2023compounding} &0.491 &45.00 &50.80& 57.60& 0.357& 26.40& 39.30& 54.50\\
GATH~\cite{wei2024enhancing} &0.463 &42.60 &47.50& 53.70& 0.344 & 25.30& 37.60 &52.70 \\
TGformer~\cite{shi2024tgformer} &0.493   &45.50 &   50.90 &56.60 &0.372 & 27.90 &  \textbf{\textcolor{blue}{41.00}}  &55.70\\
\midrule
AMIE~\cite{galarraga2013amie} &0.360 &39.10&--& 48.50& 0.230& 14.80&--&41.90 \\
AnyBURL~\cite{meilicke2019anytime} &0.454 &39.90 &--&56.20 &0.342 &25.80 &--&50.20 \\
SAFRAN~\cite{DBLP:conf/akbc/OttMS21} &0.501 &45.70 &--&58.10 &0.370 &\textbf{\textcolor{blue}{28.70}} &--&53.10 \\
\midrule
Neural LP~\cite{DBLP:conf/nips/YangYC17} &0.381 &36.80 &38.60& 40.80& 0.237& 17.30& 25.90& 36.10  \\
DRUM~\cite{DBLP:conf/nips/SadeghianADW19} &0.382 &36.90 &38.80 &41.00& 0.238& 17.40& 26.10& 36.40 \\
RLogic~\cite{DBLP:conf/kdd/ChengL0S22} &0.470 &44.30&--&53.70& 0.310& 20.30 &--&50.10 \\
RNNLogic~\cite{DBLP:conf/iclr/QuCXBT21} &0.483& 44.60 &49.70& 55.80 &0.344& 25.20 &38.00 &53.00 \\
\textsc{LatentLogic}~\cite{liu2023latentlogic} & 0.481& 45.20 &49.70 &55.30&0.320 &21.20 &32.90& 51.40\\
RNN+RotE~\cite{nandi2023simple} & \textbf{\textcolor{blue}{0.550}}& \textbf{\textcolor{blue}{51.00}} &\textbf{\textcolor{blue}{57.20}}& \textbf{\textcolor{blue}{63.50}} & 0.353& 26.50& 38.70& 52.90\\
TCRA~\cite{guo2024unified}&  0.496& 45.70 &51.10& 57.40& 0.367& 27.50& 40.30& 55.40\\
\midrule
\cellcolor{gray!50}\textsc{Tunsr} & \cellcolor{gray!50}\textbf{\textcolor{red}{0.558}}&\cellcolor{gray!50}\textbf{\textcolor{red}{51.36}}&\cellcolor{gray!50}\textbf{\textcolor{red}{58.25}}&\cellcolor{gray!50}\textbf{\textcolor{red}{65.78}}& \cellcolor{gray!50}\textbf{\textcolor{red}{0.389}} & \cellcolor{gray!50}\textbf{\textcolor{red}{28.82}} & \cellcolor{gray!50}\textbf{\textcolor{red}{41.83}} &\cellcolor{gray!50}\textbf{\textcolor{red}{57.15}}\\
\bottomrule
\end{tabular}
\label{tab_transductive}
\end{table*}

\begin{table*}[t!]
\caption{The experiment results on 12 inductive reasoning datasets.
}
\centering
\begin{tabular}
{p{0.3cm}<{\centering}rp{0.70cm}<{\centering}p{0.70cm}<{\centering}p{0.70cm}<{\centering}p{0.70cm}<{\centering}p{0.70cm}<{\centering}p{0.70cm}<{\centering}p{0.70cm}<{\centering}p{0.70cm}<{\centering}p{0.70cm}<{\centering}p{0.70cm}<{\centering}p{0.70cm}<{\centering}p{0.70cm}<{\centering}}
\toprule
&\multicolumn{1}{c}{\multirow{2}{*}{\textbf{Model}}}& \multicolumn{4}{c}{\textbf{WN18RR}}   & \multicolumn{4}{c}{\textbf{FB15k-237}}  & \multicolumn{4}{c}{\textbf{NELL-995}} \\
\cmidrule(r){3-6} \cmidrule(r){7-10} \cmidrule(r){11-14}
&\multicolumn{1}{c}{}& \textbf{V1}  & \textbf{V2} & \textbf{V3} & \textbf{V4} & \textbf{V1}  & \textbf{V2} & \textbf{V3} & \textbf{V4} & \textbf{V1}  & \textbf{V2} & \textbf{V3} & \textbf{V4} \\
\midrule
\multirow{7}{*}{\rotatebox[origin=c]{90}{\textbf{MRR}}}&GraIL~\cite{teru2020inductive} & 0.627& 0.625& 0.323& 0.553 &0.279 &0.276& 0.251& 0.227 &0.481 &0.297 &0.322& 0.262 \\
&RED-GNN~\cite{DBLP:conf/www/ZhangY22} & 0.701 & 0.690 & 0.427&0.651 &\textbf{\textcolor{blue}{0.369}}& \textbf{\textcolor{blue}{0.469}} &\textbf{\textcolor{blue}{0.445}}& \textbf{\textcolor{blue}{0.442}}& 0.637& 0.419 &\textbf{\textcolor{blue}{0.436}}& \textbf{\textcolor{blue}{0.363}} \\
&MLSAA~\cite{sun2024incorporating} & \textbf{\textcolor{blue}{0.716}}& \textbf{\textcolor{blue}{0.700}} &\textbf{\textcolor{blue}{0.448}} &\textbf{\textcolor{blue}{0.654}} &0.368 &0.457 &0.442& 0.431 &\textbf{\textcolor{blue}{0.694}}& \textbf{\textcolor{blue}{0.424}}& 0.433 &0.359 \\
\cmidrule{2-14}
&RuleN~\cite{meilicke2018fine} & 0.668& 0.645 &0.368 &0.624& 0.363 &0.433& 0.439& 0.429& 0.615 &0.385& 0.381& 0.333 \\
\cmidrule{2-14}
&Neural LP~\cite{DBLP:conf/nips/YangYC17} & 0.649& 0.635 &0.361& 0.628& 0.325& 0.389& 0.400 &0.396 &0.610& 0.361 &0.367 &0.261 \\
&DRUM~\cite{DBLP:conf/nips/SadeghianADW19} & 0.666& 0.646& 0.380 &0.627& 0.333& 0.395& 0.402 &0.410& 0.628 &0.365& 0.375& 0.273 \\
\cmidrule{2-14}
&\cellcolor{gray!50}\textsc{Tunsr} & \cellcolor{gray!50}\textbf{\textcolor{red}{0.721}} & \cellcolor{gray!50}\textbf{\textcolor{red}{0.722}} & \cellcolor{gray!50}\textbf{\textcolor{red}{0.451}} & \cellcolor{gray!50}\textbf{\textcolor{red}{0.656}}  & \cellcolor{gray!50}\textbf{\textcolor{red}{0.375}} & \cellcolor{gray!50}\textbf{\textcolor{red}{0.474}} & \cellcolor{gray!50}\textbf{\textcolor{red}{0.462}} & \cellcolor{gray!50}\textbf{\textcolor{red}{0.456}}  & \cellcolor{gray!50}\textbf{\textcolor{red}{0.746}} & \cellcolor{gray!50}\textbf{\textcolor{red}{0.427}} & \cellcolor{gray!50}\textbf{\textcolor{red}{0.455}} & \cellcolor{gray!50}\textbf{\textcolor{red}{0.387}}  \\
\midrule
\multirow{7}{*}{\rotatebox[origin=c]{90}{\textbf{Hits@1}}}&GraIL~\cite{teru2020inductive} & 55.40 &54.20& 27.80& 44.30& 20.50 &20.20& 16.50& 14.30 &42.50& 19.90 &22.40& 15.30 \\
&RED-GNN~\cite{DBLP:conf/www/ZhangY22} & 65.30 &63.30& 36.80 &60.60 &30.20 &\textbf{\textcolor{red}{38.10}} &35.10& \textbf{\textcolor{blue}{34.00}} &52.50& 31.90& \textbf{\textcolor{blue}{34.50}} &\textbf{\textcolor{blue}{25.90}} \\
&MLSAA~\cite{sun2024incorporating} & \textbf{\textcolor{blue}{66.20}}& \textbf{\textcolor{blue}{64.50}}& \textbf{\textcolor{red}{39.10}}& \textbf{\textcolor{blue}{61.20}}& 29.20 &36.60& \textbf{\textcolor{blue}{35.60}} &\textbf{\textcolor{blue}{34.00}}& \textbf{\textcolor{blue}{56.00}}& \textbf{\textcolor{red}{33.30}}& 34.30& 25.30 \\
\cmidrule{2-14}
&RuleN~\cite{meilicke2018fine} & 63.50& 61.10& 34.70 &59.20 &\textbf{\textcolor{red}{30.90}}& 34.70 &34.50& 33.80& 54.50& 30.40 &30.30& 24.80 \\ 
\cmidrule{2-14}
&Neural LP~\cite{DBLP:conf/nips/YangYC17} & 59.20& 57.50& 30.40& 58.30 &24.30& 28.60& 30.90 &28.90& 50.00& 24.90 &26.70 &13.70 \\
&DRUM~\cite{DBLP:conf/nips/SadeghianADW19} & 61.30 &59.50 &33.00& 58.60 &24.70& 28.40& 30.80& 30.90 &50.00& 27.10& 26.20& 16.30 \\
\cmidrule{2-14}
&\cellcolor{gray!50}\textsc{Tunsr} & \cellcolor{gray!50}\textbf{\textcolor{red}{66.25}} & \cellcolor{gray!50}\textbf{\textcolor{red}{66.31}} & \cellcolor{gray!50}\textbf{\textcolor{blue}{38.11}} & \cellcolor{gray!50}\textbf{\textcolor{red}{61.55}}  & \cellcolor{gray!50}\textbf{\textcolor{blue}{30.44}} & \cellcolor{gray!50}\textbf{\textcolor{blue}{37.88}} & \cellcolor{gray!50}\textbf{\textcolor{red}{37.90}} & \cellcolor{gray!50}\textbf{\textcolor{red}{36.37}}  & \cellcolor{gray!50}\textbf{\textcolor{red}{73.13}} & \cellcolor{gray!50}\textbf{\textcolor{blue}{32.67}} & \cellcolor{gray!50}\textbf{\textcolor{red}{37.13}} & \cellcolor{gray!50}\textbf{\textcolor{red}{27.30}}  \\
\midrule
\multirow{7}{*}{\rotatebox[origin=c]{90}{\textbf{Hits@10}}}&GraIL~\cite{teru2020inductive} & 76.00 &77.60 &40.90& 68.70 &42.90& 42.40 &42.40& 38.90& 56.50 &49.60& 51.80 &50.60 \\
&RED-GNN~\cite{DBLP:conf/www/ZhangY22} & 79.90 &78.00& 52.40& 72.10& 48.30& \textbf{\textcolor{blue}{62.90}}& \textbf{\textcolor{blue}{60.30}} &\textbf{\textcolor{blue}{62.10}}& 86.60&\textbf{\textcolor{blue}{60.10}}& \textbf{\textcolor{blue}{59.40}}& \textbf{\textcolor{blue}{55.60}} \\
&MLSAA~\cite{sun2024incorporating} & \textbf{\textcolor{blue}{81.10}}& \textbf{\textcolor{blue}{79.60}} &\textbf{\textcolor{blue}{54.40}}& \textbf{\textcolor{blue}{72.40}}& \textbf{\textcolor{blue}{49.00}}& 61.60& 58.90& 59.70& \textbf{\textcolor{blue}{87.80}}& 59.40& 59.20& 55.00 \\
\cmidrule{2-14}
&RuleN~\cite{meilicke2018fine} & 73.00& 69.40 &40.70 &68.10 &44.60 &59.90& 60.00& 60.50& 76.00& 51.40& 53.10& 48.40 \\
\cmidrule{2-14}
&Neural LP~\cite{DBLP:conf/nips/YangYC17} & 77.20 &74.90& 47.60& 70.60 &46.80 &58.60 &57.10& 59.30 &87.10& 56.40 &57.60 &53.90 \\
&DRUM~\cite{DBLP:conf/nips/SadeghianADW19} & 77.70& 74.70& 47.70& 70.20& 47.40& 59.50& 57.10& 59.30 &87.30& 54.00& 57.70& 53.10 \\
\cmidrule{2-14}
&\cellcolor{gray!50}\textsc{Tunsr} & \cellcolor{gray!50}\textbf{\textcolor{red}{85.87}} & \cellcolor{gray!50}\textbf{\textcolor{red}{83.98}} & \cellcolor{gray!50}\textbf{\textcolor{red}{60.76}} & \cellcolor{gray!50}\textbf{\textcolor{red}{73.28}}  & \cellcolor{gray!50}\textbf{\textcolor{red}{55.96}} & \cellcolor{gray!50}\textbf{\textcolor{red}{63.24}} & \cellcolor{gray!50}\textbf{\textcolor{red}{61.43}} & \cellcolor{gray!50}\textbf{\textcolor{red}{63.28}}  & \cellcolor{gray!50}\textbf{\textcolor{red}{88.56}} & \cellcolor{gray!50}\textbf{\textcolor{red}{62.14}} & \cellcolor{gray!50}\textbf{\textcolor{red}{61.05}} & \cellcolor{gray!50}\textbf{\textcolor{red}{58.78}}  \\

\bottomrule
\end{tabular}
\label{tab_inductive}
\end{table*}

\begin{table*}[t!]
\caption{The experiment results (Hits@10 metrics) on 12 inductive reasoning datasets with 50 negative entities for ranking.}
\centering
\begin{tabular}
{rp{2.10cm}<{\centering}p{0.70cm}<{\centering}p{0.70cm}<{\centering}p{0.70cm}<{\centering}p{0.70cm}<{\centering}p{0.70cm}<{\centering}p{0.70cm}<{\centering}p{0.70cm}<{\centering}p{0.70cm}<{\centering}p{0.70cm}<{\centering}p{0.70cm}<{\centering}p{0.70cm}<{\centering}p{0.70cm}<{\centering}}
\toprule
\multicolumn{1}{c}{\multirow{2}{*}{\textbf{Model}}}& \multicolumn{4}{c}{\textbf{WN18RR}}   & \multicolumn{4}{c}{\textbf{FB15k-237}}  & \multicolumn{4}{c}{\textbf{NELL-995}} \\
\cmidrule(r){2-5} \cmidrule(r){6-9} \cmidrule(r){10-13}
\multicolumn{1}{c}{}& \textbf{V1}  & \textbf{V2} & \textbf{V3} & \textbf{V4} & \textbf{V1}  & \textbf{V2} & \textbf{V3} & \textbf{V4} & \textbf{V1}  & \textbf{V2} & \textbf{V3} & \textbf{V4} \\
\midrule
GraIL~\cite{teru2020inductive}& 82.45& 78.68& 58.43& 73.41 & 64.15& 81.80& 82.83& 89.29 &59.50& 93.25 &91.41 &73.19\\
CoMPILE~\cite{DBLP:conf/aaai/MaiZY021}&83.60 &79.82& 60.69 &75.49 &67.64 &82.98 &84.67 &87.44&58.38 &93.87& 92.77& 75.19\\
TACT~\cite{DBLP:conf/aaai/ChenHWW21}&84.04 &81.63 &67.97 &76.56  &65.76 &83.56 &85.20 &88.69& 79.80& 88.91& 94.02 &73.78\\
\cmidrule{1-13}
RuleN~\cite{meilicke2018fine} & 80.85 &78.23& 53.39& 71.59 &49.76 &77.82& 87.69& 85.60& 53.50 &81.75& 77.26 &61.35\\
\cmidrule{1-13}
Neural LP~\cite{DBLP:conf/nips/YangYC17}& 74.37& 68.93& 46.18 &67.13 &52.92 &58.94 &52.90& 55.88 &40.78 &78.73& 82.71 &80.58\\
DRUM~\cite{DBLP:conf/nips/SadeghianADW19} & 74.37 &68.93 &46.18& 67.13 &52.92 &58.73& 52.90 &55.88 &19.42& 78.55 &82.71 &80.58\\
ConGLR~\cite{DBLP:conf/sigir/LinLXPZZZ22} &85.64 &92.93& 70.74& \textbf{\textcolor{red}{92.90}}& \textbf{\textcolor{blue}{68.29}}& \textbf{\textcolor{blue}{85.98}}& \textbf{\textcolor{blue}{88.61}}& \textbf{\textcolor{blue}{89.31}}&\textbf{\textcolor{blue}{81.07}} &\textbf{\textcolor{blue}{94.92}} &94.36& 81.61\\
SymRITa~\cite{pan2024symbolic}& \textbf{\textcolor{blue}{91.22}}& \textbf{\textcolor{blue}{88.32}}& \textbf{\textcolor{blue}{73.22}}&81.67 &74.87& 84.41& 87.11 &88.97& 64.50& 94.22& \textbf{\textcolor{blue}{95.43}}& \textbf{\textcolor{blue}{85.56}}\\
\cmidrule{1-13}
\cellcolor{gray!50}\textsc{Tunsr} & \cellcolor{gray!50}\textbf{\textcolor{red}{93.69}} & \cellcolor{gray!50}\textbf{\textcolor{red}{93.72}} & \cellcolor{gray!50}\textbf{\textcolor{red}{86.48}} & \cellcolor{gray!50}\textbf{\textcolor{blue}{89.27}}  & \cellcolor{gray!50}\textbf{\textcolor{red}{95.37}} & \cellcolor{gray!50}\textbf{\textcolor{red}{89.33}} & \cellcolor{gray!50}\textbf{\textcolor{red}{89.38}} & \cellcolor{gray!50}\textbf{\textcolor{red}{92.16}}  & \cellcolor{gray!50}\textbf{\textcolor{red}{89.05}} & \cellcolor{gray!50}\textbf{\textcolor{red}{97.91}} & \cellcolor{gray!50}\textbf{\textcolor{red}{94.69}} & \cellcolor{gray!50}\textbf{\textcolor{red}{92.63}}  \\
\bottomrule
\end{tabular}
\label{tab_inductive2}
\end{table*}

\begin{table*}[t!]
\caption{The experiment results of interpolation reasoning, including ICEWS14, ICEWS0515 and ICEWS18 datasets. 
}
\centering
\begin{tabular}
{rp{0.93cm}<{\centering}p{0.93cm}<{\centering}p{0.93cm}<{\centering}p{0.93cm}<{\centering}p{0.93cm}<{\centering}p{0.93cm}<{\centering}p{0.93cm}<{\centering}p{0.93cm}<{\centering}}
\toprule
\multicolumn{1}{c}{\multirow{2}{*}{\textbf{Model}}}& \multicolumn{4}{c}{\textbf{ICEWS14}}   & \multicolumn{4}{c}{\textbf{ICEWS0515}}   \\
\cmidrule(r){2-5} \cmidrule(r){6-9} 
\multicolumn{1}{c}{}& \textbf{MRR}  & \textbf{Hits@1} & \textbf{Hits@3} & \textbf{Hits@10} & \textbf{MRR}  & \textbf{Hits@1} & \textbf{Hits@3} & \textbf{Hits@10} \\
\midrule
TTransE~\cite{DBLP:conf/www/LeblayC18} &0.255 &7.40& --& 60.10& 27.10& 8.40& --& 61.60\\
DE-SimplE~\cite{DBLP:conf/aaai/GoelKBP20} &0.526 &41.80& 59.20& 72.50& 0.513& 39.20 &57.80& 74.80 \\
TA-DistMult~\cite{DBLP:conf/emnlp/Garcia-DuranDN18}& 0.477 &36.30 &--& 68.60& 0.474& 34.60& -- &72.80\\
ChronoR~\cite{DBLP:conf/aaai/SadeghianACW21} & 0.625 &54.70& 66.90& 77.30& 0.675& 59.60 &72.30 &82.00\\ 
TComplEx~\cite{DBLP:conf/iclr/LacroixOU20} & 0.610& 53.00& 66.00& 77.00 &0.660 &59.00& 71.00& 80.00 \\
TNTComplEx~\cite{DBLP:conf/iclr/LacroixOU20} & 0.620 &52.00 &66.00& 76.00 &0.670 &59.00 &71.00& 81.00 \\
TeLM~\cite{DBLP:conf/naacl/XuCNL21} &0.625& 54.50& 67.30& 77.40 &0.678 &59.90& 72.80 &82.30 \\
BoxTE~\cite{DBLP:conf/aaai/MessnerAC22} & 0.613 &52.80& 66.40& 76.30& 0.667& 58.20& 71.90& 82.00 \\
RotateQVS~\cite{DBLP:conf/acl/ChenWLL22}& 0.591& 50.70& 64.20& 75.04& 0.633& 52.90& 70.90& 81.30\\
TeAST~\cite{DBLP:conf/acl/LiSG23}& \textcolor{blue}{\textbf{0.637}} &\textcolor{blue}{\textbf{56.00}} &\textcolor{blue}{\textbf{68.20}}& \textcolor{blue}{\textbf{78.20}}& \textcolor{blue}{\textbf{0.683}} &\textcolor{red}{\textbf{60.40}}& \textcolor{blue}{\textbf{73.20}} &\textcolor{blue}{\textbf{82.90}}\\
\midrule
\cellcolor{gray!50}\textsc{Tunsr} & \cellcolor{gray!50}\textcolor{red}{\textbf{0.648}}&\cellcolor{gray!50}\textcolor{red}{\textbf{56.21}}&\cellcolor{gray!50}\textcolor{red}{\textbf{69.61}}&\cellcolor{gray!50}\textcolor{red}{\textbf{80.16}}&\cellcolor{gray!50}\textcolor{red}{\textbf{0.705}}&\cellcolor{gray!50}\textcolor{blue}{\textbf{59.89}}&\cellcolor{gray!50}\textcolor{red}{\textbf{74.67}}&\cellcolor{gray!50}\textcolor{red}{\textbf{83.95}}\\
\bottomrule
\end{tabular}
\label{tab_interpolation}
\end{table*}

\begin{table*}[t!]
\caption{The experiment results of extrapolation reasoning, including ICEWS14, ICEWS0515, and ICEWS18 datasets.
}
\centering
\begin{tabular}
{rp{0.70cm}<{\centering}p{0.70cm}<{\centering}p{0.70cm}<{\centering}p{0.70cm}<{\centering}p{0.70cm}<{\centering}p{0.70cm}<{\centering}p{0.70cm}<{\centering}p{0.70cm}<{\centering}p{0.70cm}<{\centering}p{0.70cm}<{\centering}p{0.70cm}<{\centering}p{0.70cm}<{\centering}}
\toprule
\multicolumn{1}{c}{\multirow{2}{*}{\textbf{Model}}}& \multicolumn{4}{c}{\textbf{ICEWS14}}   & \multicolumn{4}{c}{\textbf{ICEWS0515}}  & \multicolumn{4}{c}{\textbf{ICEWS18}}    \\
\cmidrule(r){2-5} \cmidrule(r){6-9} \cmidrule(r){10-13}
\multicolumn{1}{c}{}& \textbf{MRR}  & \textbf{Hits@1} & \textbf{Hits@3} & \textbf{Hits@10} & \textbf{MRR}  & \textbf{Hits@1} & \textbf{Hits@3} & \textbf{Hits@10} & \textbf{MRR}  & \textbf{Hits@1} & \textbf{Hits@3} & \textbf{Hits@10} \\
\midrule
TransE~\cite{DBLP:conf/nips/BordesUGWY13} & 0.224 & 13.36 & 25.63 & 41.23  & 0.225 & 13.05 & 25.61 & 42.05  & 0.122 & 5.84  & 12.81 & 25.10  \\
DistMult~\cite{DBLP:journals/corr/YangYHGD14a} & 0.276 & 18.16 & 31.15 & 46.96  & 0.287 & 19.33 & 32.19 & 47.54  & 0.107 & 4.52  & 10.33 & 21.25  \\
ComplEx~\cite{DBLP:conf/icml/TrouillonWRGB16} & 0.308 & 21.51 & 34.48 & 49.58  & 0.316 & 21.44 & 35.74 & 52.04  & 0.210 & 11.87 & 23.47 & 39.87  \\
TTransE~\cite{DBLP:conf/www/LeblayC18} & 0.134 & 3.11  & 17.32 & 34.55  & 0.157 & 5.00  & 19.72 & 38.02  & 0.083 & 1.92  & 8.56  & 21.89  \\
TA-DistMult~\cite{DBLP:conf/emnlp/Garcia-DuranDN18} & 0.264 & 17.09 & 30.22 & 45.41  & 0.243 & 14.58 & 27.92 & 44.21  & 0.167 & 8.61  & 18.41 & 33.59  \\
TA-TransE~\cite{DBLP:conf/emnlp/Garcia-DuranDN18} & 0.174 & 0.00  & 29.19 & 47.41  & 0.193 & 1.81  & 31.34 & 50.33  & 0.125 & 0.01  & 17.92 & 37.38  \\
DE-SimplE~\cite{DBLP:conf/aaai/GoelKBP20} & 0.326 & 24.43 & 35.69 & 49.11  & 0.350 & 25.91 & 38.99 & 52.75  & 0.193 & 11.53 & 21.86 & 34.80  \\
TNTComplEx~\cite{DBLP:conf/iclr/LacroixOU20} & 0.321 & 23.35 & 36.03 & 49.13  & 0.275 & 19.52 & 30.80 & 42.86  & 0.212 & 13.28 & 24.02 & 36.91  \\
RE-Net~\cite{DBLP:conf/emnlp/JinQJR20} & 0.382 & 28.68 & 41.34 & 54.52  & 0.429 & 31.26 & 46.85 & 63.47  & 0.288 & 19.05 & 32.44 & 47.51  \\
CyGNet~\cite{DBLP:conf/aaai/ZhuCFCZ21} & 0.327 & 23.69 & 36.31 & 50.67  & 0.349 & 25.67 & 39.09 & 52.94  & 0.249 & 15.90 & 28.28 & 42.61  \\
\midrule
AnyBURL~\cite{meilicke2019anytime} & 0.296 & 21.26 & 33.33 & 46.73  & 0.320 & 23.72 & 35.45 & 50.46  & 0.227 & 15.10 & 25.44 & 38.91  \\
TLogic~\cite{DBLP:conf/aaai/LiuMHJT22} & 0.430 & 33.56 & 48.27 & 61.23  & 0.469 & 36.21 & 53.13 & 67.43  & 0.298 & 20.54 & 33.95 & 48.53  \\
TR-Rules~\cite{DBLP:conf/emnlp/LiELSYSWL23} & 0.433 &33.96 &48.55 &61.17 &0.476 &37.06 &53.80 &67.57 &0.304 &21.10 &34.58 &48.92  \\
\midrule
xERTE~\cite{DBLP:conf/iclr/HanCMT21} & 0.407 & 32.70 & 45.67 & 57.30  & 0.466 & 37.84 & 52.31 & 63.92  & 0.293 & 21.03 & 33.51 & 46.48  \\
TITer~\cite{DBLP:conf/emnlp/SunZMH021} & 0.417 & 32.74 & 46.46 & 58.44  & -- & -- & -- & -- & 0.299 & \textcolor{blue}{\textbf{22.05}} & 33.46 & 44.83  \\
TECHS~\cite{DBLP:conf/acl/LinL0XC23} & 0.438&\textcolor{blue}{\textbf{34.59}}&\textcolor{blue}{\textbf{49.36}}&61.95& \textcolor{blue}{\textbf{0.483}} & \textcolor{red}{\textbf{38.34}} & \textcolor{blue}{\textbf{54.69}}&\textcolor{blue}{\textbf{68.92}}&0.308&21.81&35.39&49.82\\
INFER~\cite{liinfer}& \textcolor{blue}{\textbf{0.441}}& 34.52 &48.92 &\textcolor{blue}{\textbf{62.14}}& 0.483 &37.61& 54.30& 68.52 &\textcolor{blue}{\textbf{0.317}} &21.94& \textcolor{blue}{\textbf{35.64}}& \textcolor{blue}{\textbf{50.88}}\\
\midrule
\cellcolor{gray!50}\textsc{Tunsr} & \cellcolor{gray!50}\textcolor{red}{\textbf{0.447}}&\cellcolor{gray!50}\textcolor{red}{\textbf{35.16}}&\cellcolor{gray!50}\textcolor{red}{\textbf{50.39}}&\cellcolor{gray!50}\textcolor{red}{\textbf{63.32}}&\cellcolor{gray!50}\textcolor{red}{\textbf{0.491}} & \cellcolor{gray!50}\textcolor{blue}{\textbf{38.31}} & \cellcolor{gray!50}\textcolor{red}{\textbf{55.67}}&\cellcolor{gray!50}\textcolor{red}{\textbf{69.88}}&\cellcolor{gray!50}\textcolor{red}{\textbf{0.321}}&\cellcolor{gray!50}\textcolor{red}{\textbf{22.99}}&\cellcolor{gray!50}\textcolor{red}{\textbf{36.68}}&\cellcolor{gray!50}\textcolor{red}{\textbf{51.08}}\\
\bottomrule
\end{tabular}
\label{tab_extrapolation}
\end{table*}

The experiments on transductive, inductive, interpolation, and extrapolation reasoning are carried out to evaluate the performance.
The results are shown in Tables~\ref{tab_transductive}, \ref{tab_inductive}, \ref{tab_interpolation} and \ref{tab_extrapolation}, respectively.
It can be observed that our model has performance advantages over neural, symbolic, and neurosymbolic methods.

Specifically, from Table~\ref{tab_transductive} of transductive reasoning, it is observed that \textsc{Tunsr} achieves the optimal performance.
Compared with advanced neural methods, \textsc{Tunsr} shows performance advantages.
For example, it improves the Hits@10 values of the two datasets by 8.78\%, 16.78\%, 8.48\%, 8.68\%, 9.08\%, 3.38\%, 8.18\%, 12.08\%
and 4.45\%, 15.25\%, 3.05\%, 1.15\%, 1.95\%, 1.35\%, 2.65\%, 4.45\%
compared with TransE, DistMult, UltraE, ComplEx-DURA, AutoBLM, RED-GNN, CompoundE and GATH model.
Moreover, compared with symbolic and neurosymbolic methods, the advantages of the \textsc{Tunsr} are more obvious.
For symbolic methods (AMIE, AnyBURL, and SAFRAN), the average achievements of MRR, Hits@1, and Hits@10 values are 0.119, 9.79\%, 11.51\% and 0.075, 5.72\%, 8.75\% on two datasets.
For advanced neurosymbolic RNNLogic, \textsc{LatentLogic} and RNN+RotE, \textsc{Tunsr} also shows performance advantages, achieving 9.98\%, 10.48\%, 2.28\% and 4.15\%, 5.75\%, 4.25\% of Hits@10 improvements on two datasets.

For inductive reasoning, \textsc{Tunsr} also has the performance advantage compared with all neural, symbolic, and neurosymbolic methods as Table~\ref{tab_inductive} shows, especially on WN18RR v1, WN18RR v2, WN18RR v3, FB15k-237 v1, and NELL-995 v1 datasets.
Specifically, \textsc{Tunsr} is better than neural methods GraIL, MLSAA, and RED-GNN. 
Compared with the latter, it achieves 5.97\%, 5.98\%, 8.36\%, 1.18\%, 7.66\%, 0.34\%, 1.13\%, 1.18\%, 1.96\%, 2.04\%, 1.65\%, and 3.18\% improvements on the His@10 metric, reaching an average improvement of 3.39\%.
For symbolic and neurosymbolic methods (RuleN, Neural LP, and DRUM), \textsc{Tunsr} has greater performance advantages.
For example, compared with DRUM, \textsc{Tunsr} has achieved an average improvement of 0.069, 8.19\%, and 6.05\% on MRR, Hits@1, and Hits@10 metrics, respectively.
Besides, for equal comparison with CoMPILE~\cite{DBLP:conf/aaai/MaiZY021}, TACT~\cite{DBLP:conf/aaai/ChenHWW21}, ConGLR~\cite{DBLP:conf/sigir/LinLXPZZZ22}, and SymRITa~\cite{pan2024symbolic}, we carry out the evaluation under their setting which introduces 50 negative entities (rather than all entities) for ranking for each query. The results are shown in Table~\ref{tab_inductive2}. These results also verify the superiority of our model.

\begin{figure*}[t]
  \centering
  \begin{minipage}[t]{0.24\linewidth}
    \large
    \centering
    \includegraphics[scale=0.15]{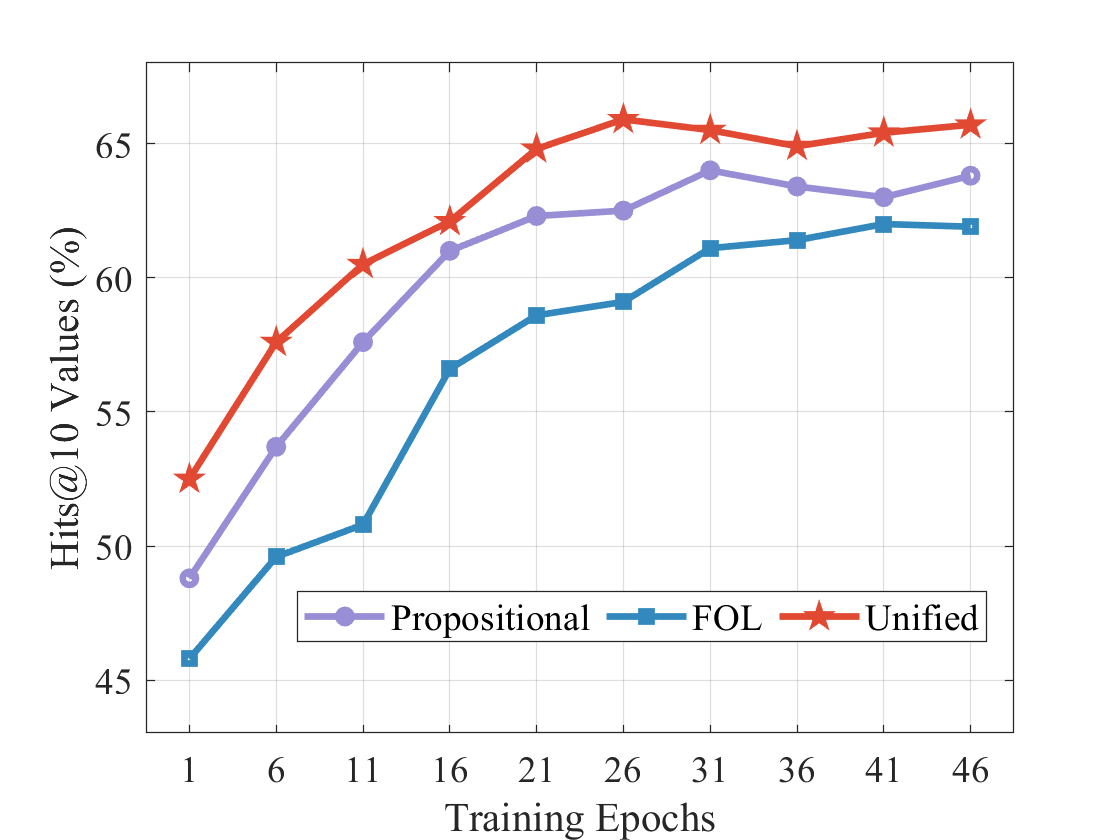}
    \subcaption{WN18RR of SKG$_{T}$.}
    \label{fig_ablation1}
  \end{minipage}
  \begin{minipage}[t]{0.24\linewidth}
    \large
    \centering
    \includegraphics[scale=0.15]{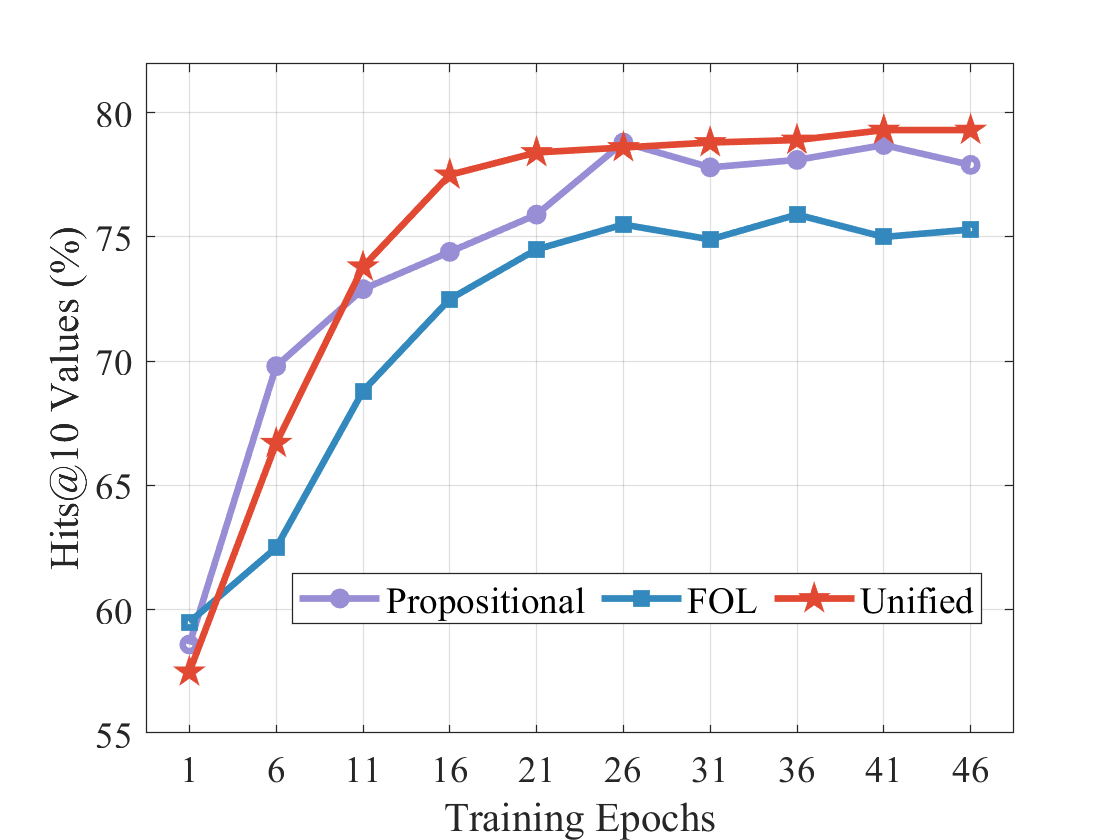}
    \subcaption{ICEWS14 of TKG$_{I}$.}
    \label{fig_ablation2}
  \end{minipage}
  \begin{minipage}[t]{0.24\linewidth}
    \large
    \centering
    \includegraphics[scale=0.15]{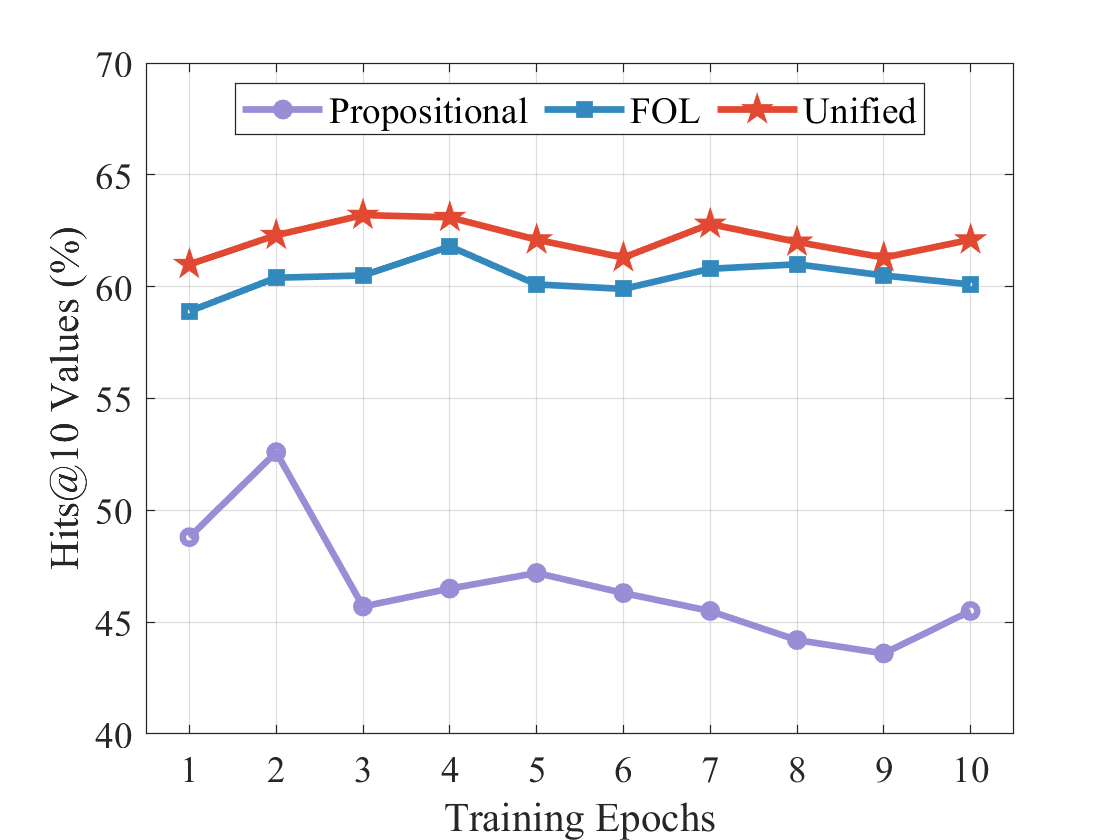}
    \subcaption{ICEWS14 of TKG$_{E}$.}
    \label{fig_ablation3}
  \end{minipage}
  \begin{minipage}[t]{0.24\linewidth}
    \large
    \centering
    \includegraphics[scale=0.15]{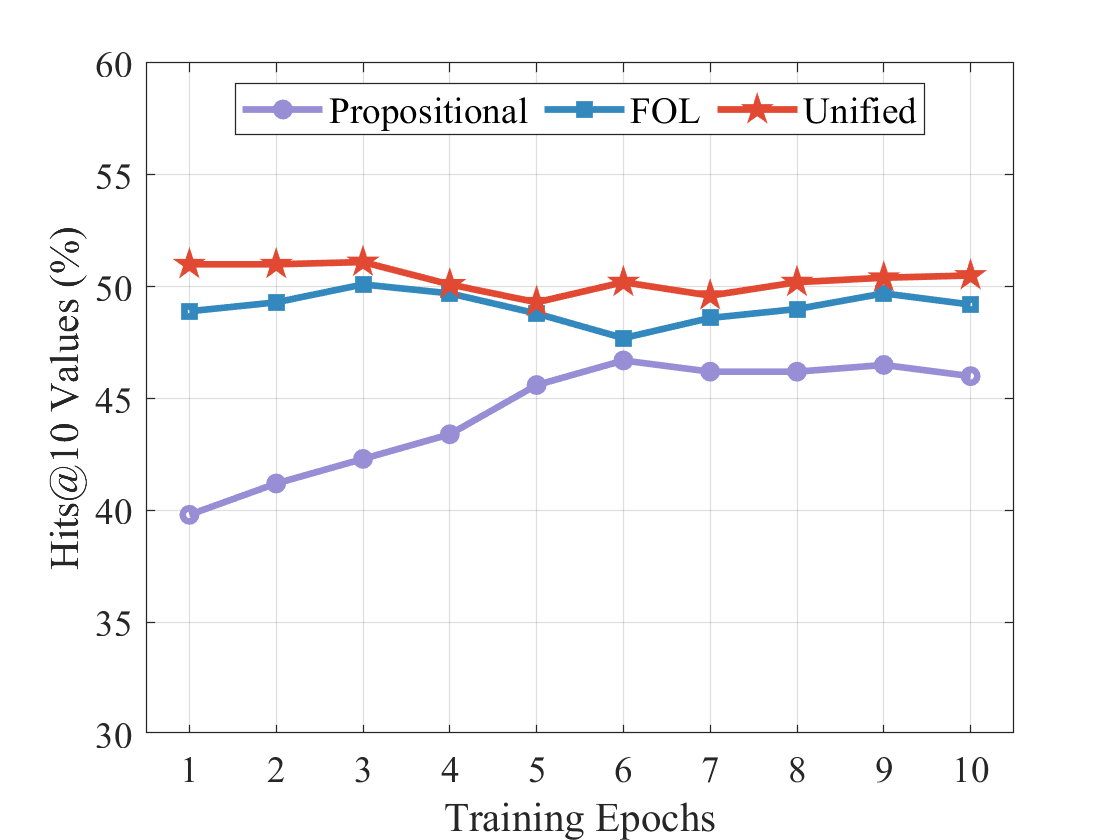}
    \subcaption{ICEWS18 of TKG$_{E}$.}
    \label{fig_ablation4}
  \end{minipage}
  \caption{The impacts of propositional and FOL reasoning on transductive, interpolation, and extrapolation scenarios. It is generally observed that the unified model has a better performance compared with the single propositional or FOL setting, demonstrating the validity and rationality of the unified mechanism in \textsc{Tunsr}.}
  \label{fig_ablation}
\end{figure*}

For interpolation reasoning in Table~\ref{tab_interpolation}, the performance of \textsc{Tunsr} surpasses that of mainstream neural reasoning methods.
It achieves optimal results on seven out of eight metrics.
Compared with TNTComplEx of the previous classic tensor-decomposition method, the improvement on each metric is 0.028, 4.21\%, 3.61\%, 4.16\%, 0.035, 0.89\%, 3.67\%, and 2.95\%, respectively.
Moreover, compared with the state-of-the-art model TeAST that encodes temporal knowledge graph embeddings via the archimedean spiral timeline,
\textsc{Tunsr} also has 0.011, 0.21\%, 1.41\%, 1.96\%, 0.022, -0.51\%, 1.47\%, and 1.05\% performance advantages (only slightly smaller on Hits@1 metric of ICEWS0515 dataset).

As Table~\ref{tab_extrapolation} shows for extrapolation reasoning, \textsc{Tunsr} also performs better.
Compared with 10 neural reasoning methods, \textsc{Tunsr} has obvious performance advantages.
For instance, it achieves 14.19\%, 27.02\%, and 14,17\% Hits@10 improvement on three datasets against the tensor-decomposition method TNTComplEx. Additionally, \textsc{Tunsr} outperforms symbolic rule-based methods, i.e., AnyBURL, TLogic, and TR-Rules, achieving average improvements of 0.061,	5.57\%,	7.01\%,	6.94\%,	0.069, 5.98\%, 8.21\%, 8.06\%, 0.045, 4.08\%, 5.36\%, and 5.63\% on all 12 evaluation metrics.
Moreover, \textsc{Tunsr} excels three neurosymbolic methods (xERTE, TITer and INFER) across all datasets.
Furthermore, compared with the previous study TECHS, \textsc{Tunsr} also has the performance boost, which shows 1.37\%, 0.96\%, and 1.26\% Hits@10 metric gains.

In summary, the experimental results on four reasoning scenarios demonstrate the effectiveness and superiority of the proposed unified framework \textsc{Tunsr}.
It shows the rationality of the unified mechanism at both the methodological and application perspectives and verifies the tremendous potential for future KG reasoning frameworks.

\subsection{Ablation Studies (RQ2)}

To explore the impacts of propositional and FOL parts on KG reasoning performance,
we carry out ablation studies on transductive (WN18RR), interpolation (ICEWS14), and extrapolation (ICEWS14 and ICEWS18) scenarios in Figure~\ref{fig_ablation}.
As inductive reasoning is entity-independent, we only conduct experiments using FOL reasoning for it.
In each line chart, we depict the performance trends associated with propositional, FOL, and unified reasoning throughout the training epochs.
In the propositional/FOL setting, we set $\lambda$ in the Eq. (\ref{eq14}) as 0/1, indicating the model only uses propositional/FOL reasoning to get the answer. 
In the unified setting, the value of $\lambda$ is the dynamic learned by embeddings.
From the figure, it is generally observed that the unified setting has a better performance compared with the single propositional or FOL setting.
It is noteworthy that propositional and FOL display unique characteristics when applied to diverse datasets.
For transductive and interpolation reasoning (Figures~\ref{fig_ablation1} and ~\ref{fig_ablation2}), the performance of propositional reasoning consistently surpasses that of FOL, despite both exhibiting continuous improvement throughout the training process.
However, it is contrary to the results on the extrapolation scenario (Figures~\ref{fig_ablation3} and ~\ref{fig_ablation4}),
where FOL reasoning has performance advantages.
It is noted that propositional reasoning performs well in ICEWS18 while badly in ICEWS14 under the extrapolation setting.
This may be caused by the structural differences between ICEWS14 and ICEWS18. Compared with ICEWS14, the graph structure of ICEWS18 is notably denser (8.94 vs. 16.19 in node degree).
So propositional reasoning in ICEWS18 can capture more comprehensive pattern semantics and exhibit robust generalization in testing scenarios.
These observations indicate that propositional and FOL reasoning emphasizes distinct aspects of knowledge.
Thus, combining them allows for the synergistic exploitation of their respective strengths, resulting in an enhanced overall effect.

\begin{figure*}[t]
  \centering
  \begin{minipage}[t]{0.24\linewidth}
    \large
    \centering
    \includegraphics[scale=0.15]{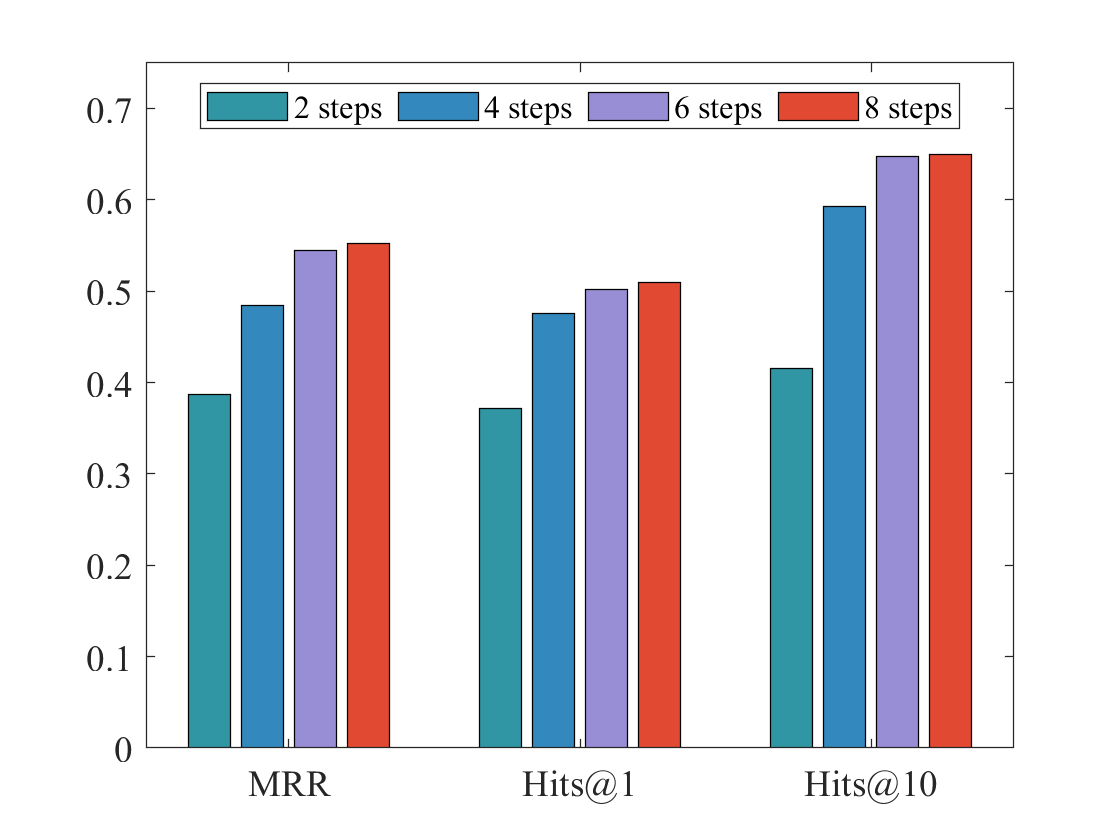}
    \subcaption{WN18RR of SKG$_{T}$.}
    \label{fig_length1}
  \end{minipage}
  \begin{minipage}[t]{0.24\linewidth}
    \large
    \centering
    \includegraphics[scale=0.15]{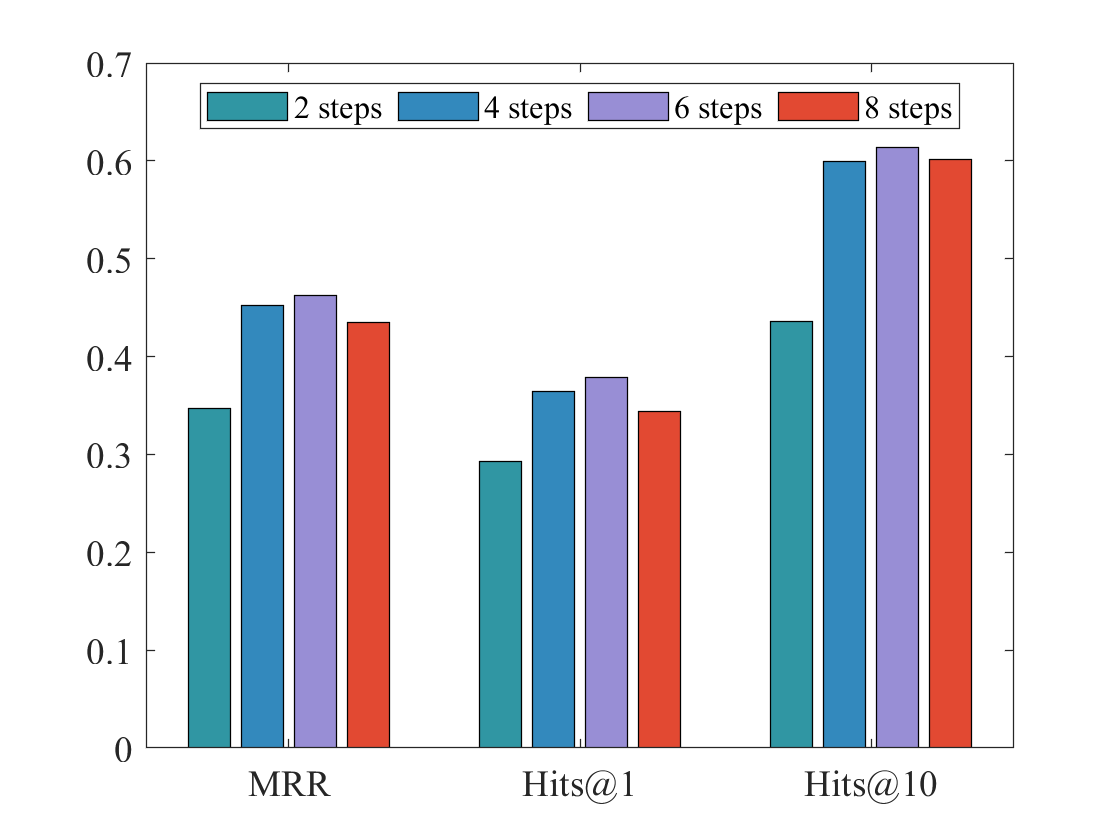}
    \subcaption{FB15k-237 v3 of SKG$_{I}$.}
    \label{fig_length2}
  \end{minipage}
  \begin{minipage}[t]{0.24\linewidth}
    \large
    \centering
    \includegraphics[scale=0.15]{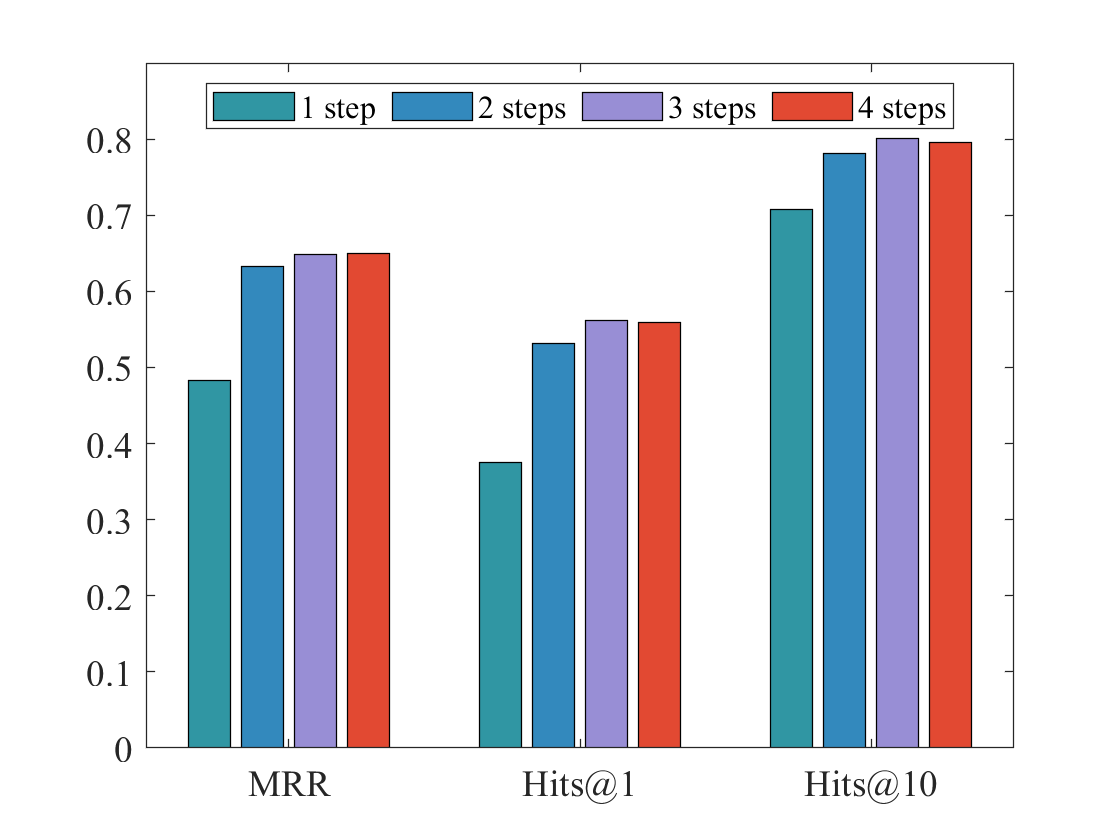}
    \subcaption{ICEWS14 of TKG$_{I}$.}
    \label{fig_length3}
  \end{minipage}
  \begin{minipage}[t]{0.24\linewidth}
    \large
    \centering
    \includegraphics[scale=0.15]{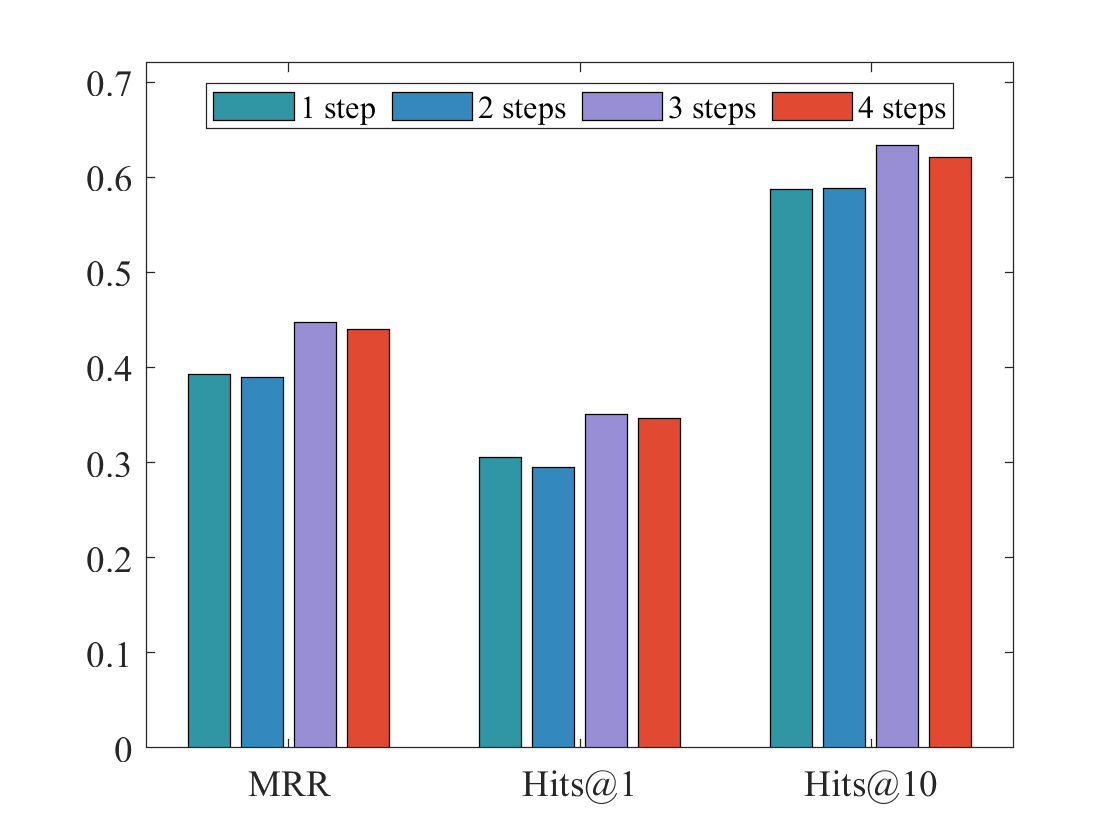}
    \subcaption{ICEWS14 of TKG$_{E}$.}
    \label{fig_length4}
  \end{minipage}
  \caption{The impacts of reasoning iterations which correspond to the length of the reasoning rules. It is evident that choosing the appropriate value is crucial for obtaining accurate reasoning results.}
  \label{fig_length}
\end{figure*}

\begin{figure*}[t]
  \centering
  \begin{minipage}[t]{0.24\linewidth}
    \large
    \centering
    \includegraphics[scale=0.14]{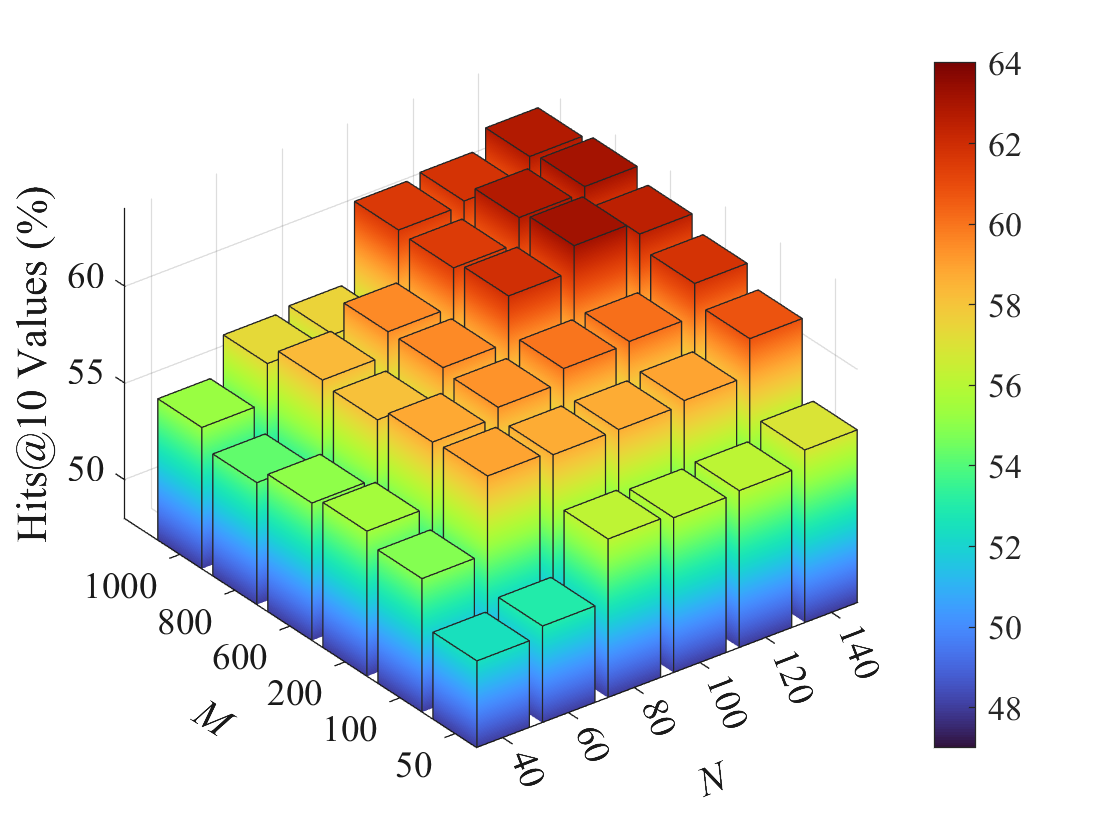}
    \subcaption{Performance on ICEWS14.}
    \label{fig_mn1}
  \end{minipage}
  \begin{minipage}[t]{0.24\linewidth}
    \large
    \centering
    \includegraphics[scale=0.14]{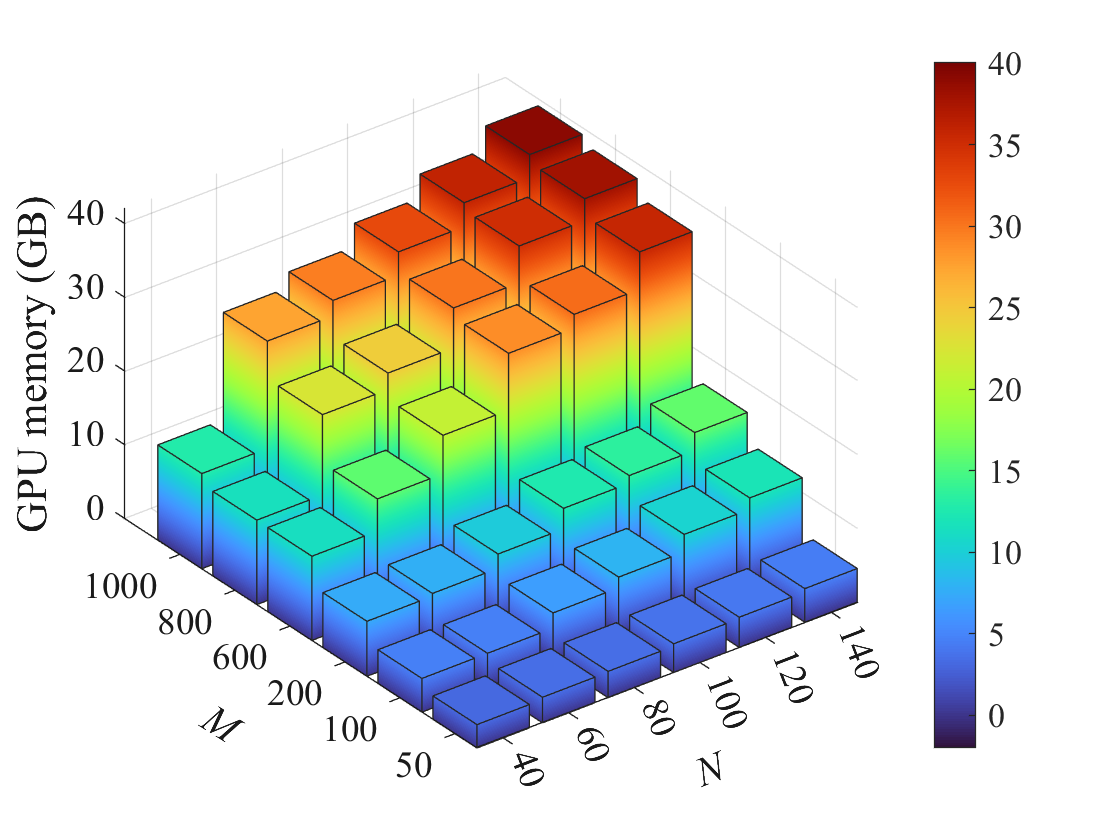}
    \subcaption{Space on ICEWS14.}
    \label{fig_mn2}
  \end{minipage}
  \begin{minipage}[t]{0.24\linewidth}
    \large
    \centering
    \includegraphics[scale=0.14]{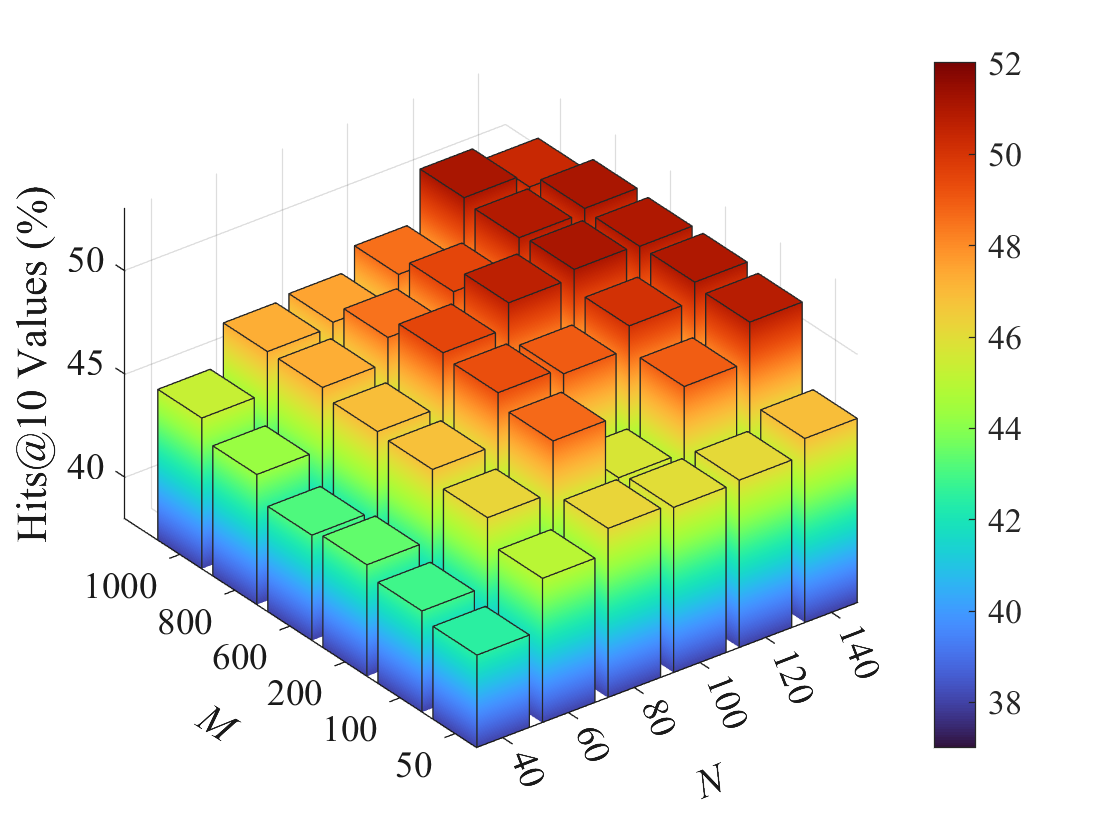}
    \subcaption{Performance on ICEWS18.}
    \label{fig_mn3}
  \end{minipage}
  \begin{minipage}[t]{0.24\linewidth}
    \large
    \centering
    \includegraphics[scale=0.14]{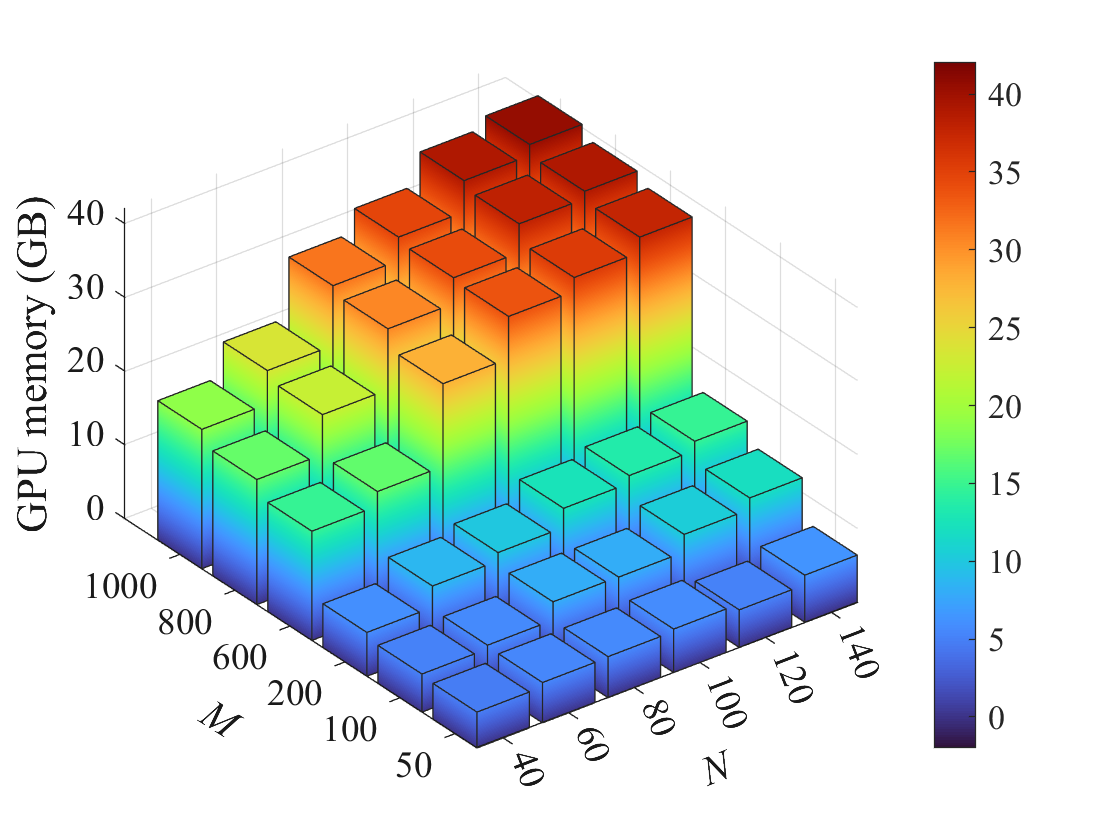}
    \subcaption{Space on ICEWS18.}
    \label{fig_mn4}
  \end{minipage}
  \caption{The impacts of sampling in the reasoning process. Performance and GPU space usage with batch size 64. Large values of \emph{M} and \emph{N} can achieve excellent performance at the cost of increased space requirements.}
  \label{fig_mn}
\end{figure*}

\subsection{Hyperparameter Analysis (RQ3)}

We run our model with different hyperparameters to explore weight impacts in Figures~\ref{fig_length} and~\ref{fig_mn}.
Specifically, Figure~\ref{fig_length} illustrates the performance variation with different reasoning iterations, i.e., the length of the reasoning rules.
At WN18RR and FB15k-237 v3 datasets of transductive and inductive settings,
experiments on rule lengths of 2, 4, 6, and 8 are carried out as illustrated in Figures~\ref{fig_length1} and~\ref{fig_length2}.
It is observed that the performance generally improves with the iteration increasing from 2 to 6.
When the rule length continues to increase, the inference performance changes little or decreases slightly.
The same phenomenon can also be observed in Figures~\ref{fig_length3} and \ref{fig_length4},
which corresponds to interpolation and extrapolation reasoning on the ICEWS14 dataset.
The rule length ranges from 1 to 4, where the model performance typically displays an initial improvement, followed by a tendency to stabilize or exhibit a marginal decline.
This phenomenon occurs due to the heightened rule length,
which amplifies the modeling capability while potentially introducing noise into the reasoning process.
Therefore, an appropriate value of rule length (reasoning iteration) is significant for KG reasoning.

We also explore the impacts of hyperparameters \emph{M} for node sampling and \emph{N} for edge selection on ICEWS14 and ICEWS18 datasets of extrapolation reasoning.
The results are shown in Figure~\ref{fig_mn}.
For each dataset, we show the reasoning performance (Hits@10 metric) and utilized space (GPU memory) in detail, with the \emph{M} varies in \{50, 100, 200, 600, 800, 1000\} while \emph{N} varies in \{40, 60, 80, 100, 120, 140\}.
It is evident that opting for smaller values results in a significant decline in performance. This decline can be attributed to the inadequate number of nodes and edges, which respectively contribute to insufficient and unstable training.
Furthermore, as \emph{M} surpasses 120, the marginal gains become smaller or even lead to performance degradation.
Additionally, when \emph{M} and \emph{N} are increased, the GPU memory utilization of the model experiences rapid growth, as depicted in Figure~\ref{fig_mn2} and~\ref{fig_mn4}, with a particularly pronounced effect on \emph{M}.

\begin{table*}[t!]
\centering
\caption{Some reasoning cases in transductive, interpolation, and extrapolation scenarios, where both propositional reasoning and learned FOL rules are displayed.
``${-1}$'' denotes the reverse of a specific relation and textual descriptions of some relations are simplified. Values in orange rectangles represent propositional attentions and relations marked with red in FOL rules represent the target relation to be predicted.}
\begin{tabular}{cc}
\toprule
Propositional Reasoning& FOL Rules\\
\midrule
\hspace{-0.5cm}\begin{minipage}{0.35\textwidth}
\centering
\includegraphics[scale=0.56]{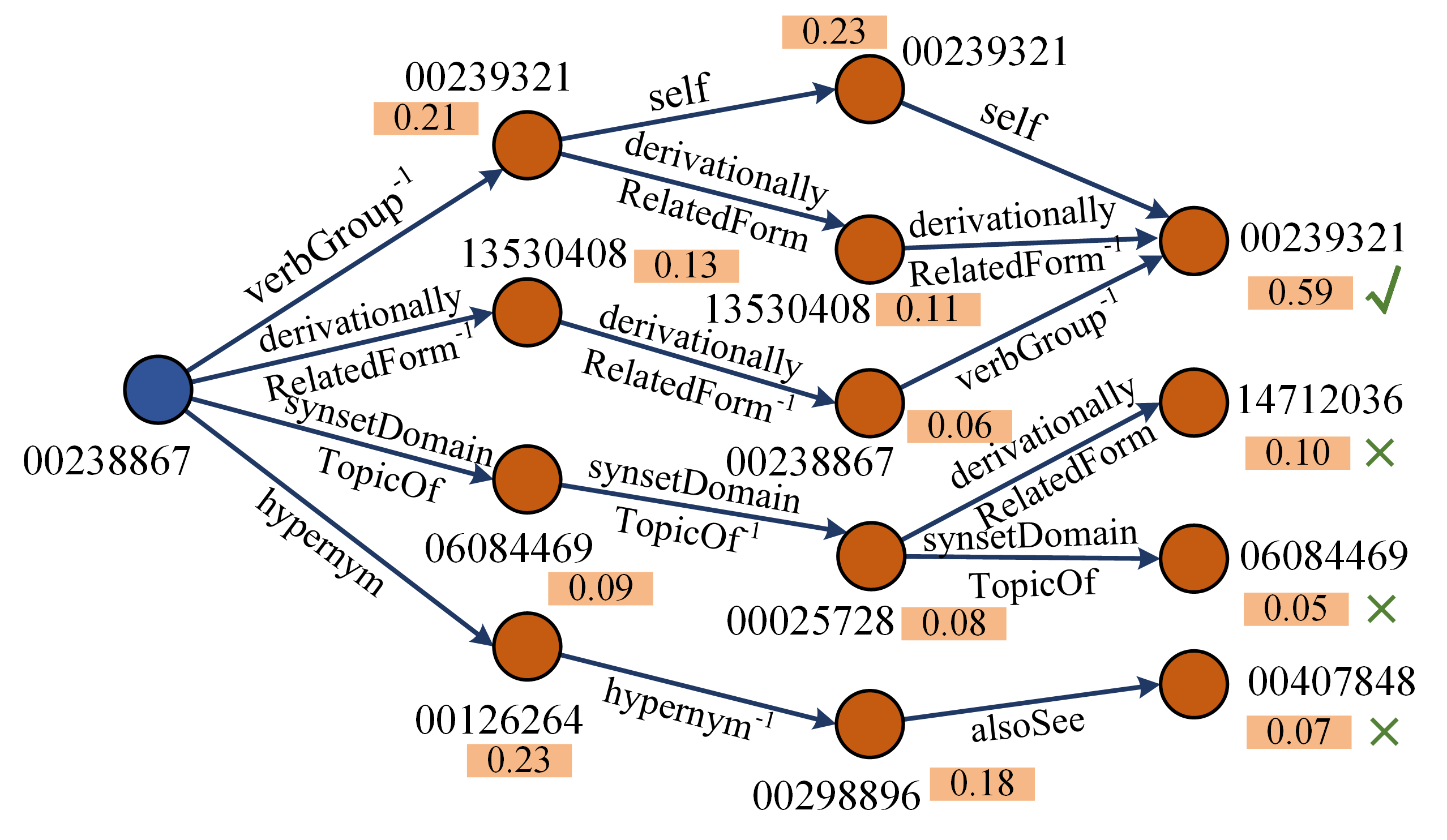}
\end{minipage}
&
\begin{minipage}{0.6\textwidth}
\begin{enumerate}[label={[\arabic{enumi}]},leftmargin=*, itemsep=1pt]
\item 0.21 \emph{\textcolor{red}{verbGroup}}$^{-1}$\emph{(X,Z)} $\rightarrow$ \emph{\textcolor{red}{verbGroup}(X,Z)}
\item 0.32 \emph{\textcolor{red}{verbGroup}}$^{-1}$\emph{(X,Y$_1$)}$\land$\emph{derivationallyRelatedForm}\emph{(Y$_1$,Y$_2$)}$\land$\\\emph{derivationallyRelatedForm}$^{-1}$\emph{(Y$_2$,Z)} $\rightarrow$ \emph{\textcolor{red}{verbGroup}(X,Z)}
\item 0.07 \emph{derivationallyRelatedForm}$^{-1}$\emph{(X,Y$_1$)}$\land$\emph{derivationallyRelatedForm}$^{-1}$\emph{(Y$_1$,Y$_2$)}$\land$\\\emph{\textcolor{red}{verbGroup}}$^{-1}$\emph{(Y$_2$,Z)} $\rightarrow$ \emph{\textcolor{red}{verbGroup}(X,Z)}
\item 0.05 \emph{synsetDomainTopicOf}\emph{(X,Y$_1$)}$\land$\emph{synsetDomainTopicOf}$^{-1}$\emph{(Y$_1$,Y$_2$)}$\land$\\\emph{derivationallyRelatedForm}\emph{(Y$_2$,Z)}$\rightarrow$\emph{\textcolor{red}{verbGroup}(X,Z)}
\item 0.18 \emph{hypernym}\emph{(X,Y$_1$)}$\land$\emph{hypernym}$^{-1}$\emph{(Y$_1$,Y$_2$)}$\land$\emph{alsoSee}\emph{(Y$_2$,Z)}$\rightarrow$\emph{\textcolor{red}{verbGroup}(X,Z)}
\end{enumerate}
\end{minipage}\\
\multicolumn{2}{c}{\cellcolor{gray!30}Transductive reasoning: query (\emph{00238867}, \emph{verbGroup}, ?) in WN18RR}\\

\midrule
\hspace{-0.5cm}\begin{minipage}{0.35\textwidth}
\centering
\includegraphics[scale=0.56]{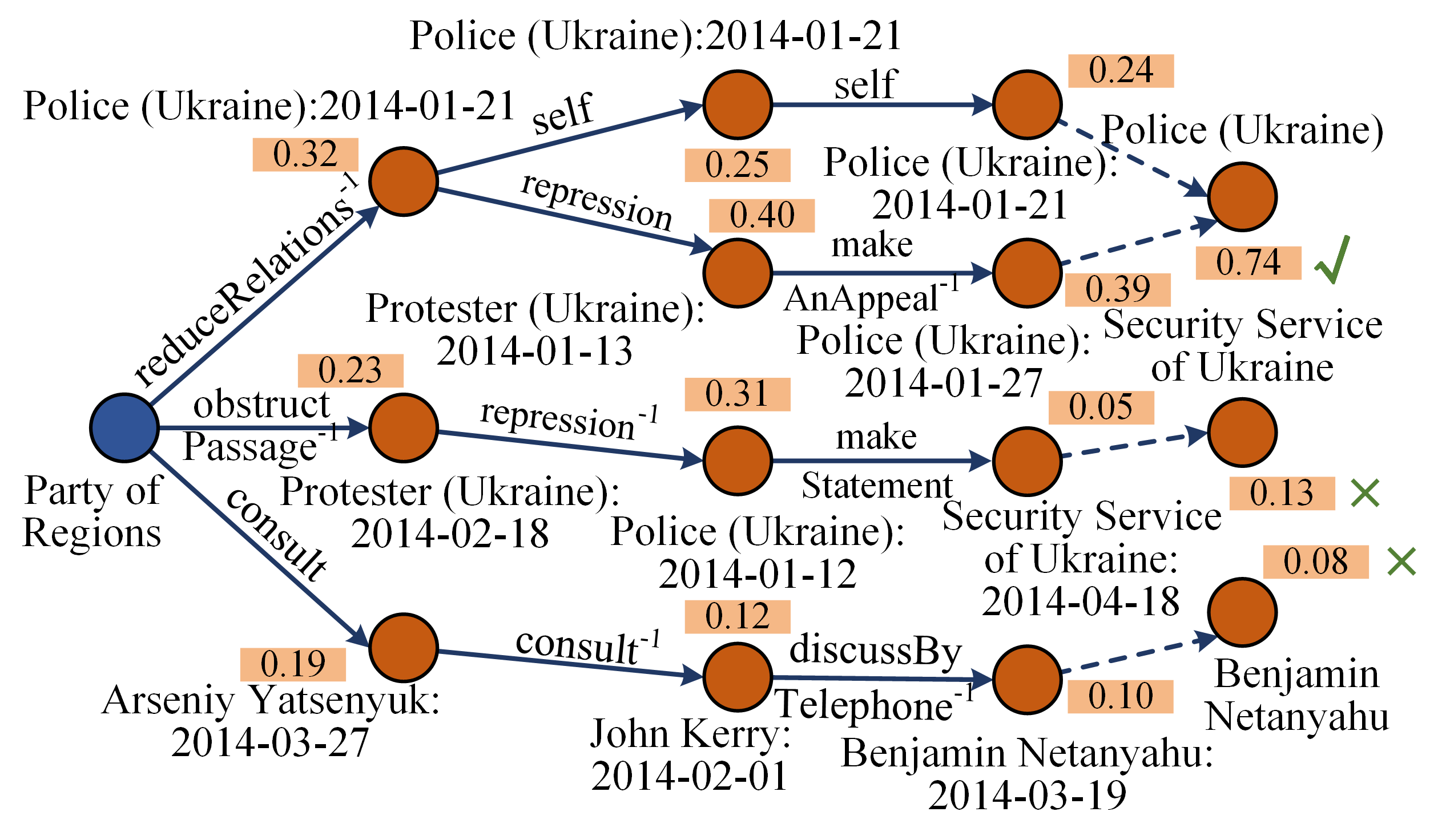}
\end{minipage}
&
\begin{minipage}{0.6\textwidth}
\begin{enumerate}[label={[\arabic{enumi}]},leftmargin=*, itemsep=1pt]
\item 0.46 \emph{reduceRelations}$^{-1}$\emph{(X,Z)}$:t_1$ $\rightarrow$\emph{\textcolor{red}{makeAnAppeal}(X,Z)}$:t$
\item 0.19 \emph{reduceRelations}$^{-1}$\emph{(X,Y$_1$)}$:t_1$$\land$\emph{repression}\emph{(Y$_1$,Y$_2$)}$:t_2$$\land$\emph{\textcolor{red}{makeAnAppeal}}$^{-1}$\emph{(Y$_2$,Z)}$:t_3$\\$\rightarrow$ \emph{\textcolor{red}{makeAnAppeal}(X,Z)}$:t$
\item 0.14 \emph{obstructPassage}$^{-1}$\emph{(X,Y$_1$)}$:t_1$$\land$\emph{repression}$^{-1}$\emph{(Y$_1$,Y$_2$)}$:t_2$$\land$\emph{makeStatement}\emph{(Y$_2$,Z)}$:t_3$\\$\rightarrow$\emph{\textcolor{red}{makeAnAppeal}(X,Z)}$:t$
\item 0.12 \emph{consult}\emph{(X,Y$_1$)}$:t_1$$\land$\emph{consult}$^{-1}$\emph{(Y$_1$,Y$_2$)}$:t_2$$\land$\emph{discussByTelephone}$^{-1}$\emph{(Y$_2$,Z)}$:t_3$\\$\rightarrow$\emph{\textcolor{red}{makeAnAppeal}(X,Z)}$:t$
\end{enumerate}
\end{minipage}\\
\multicolumn{2}{c}{\cellcolor{gray!30}Interpolation reasoning: query (\emph{Party of Regions}, \emph{makeAnAppeal}, ?, \emph{2014-05-15}) in ICEWS14}\\

\midrule
\hspace{-0.5cm}\begin{minipage}{0.35\textwidth}
\centering
\includegraphics[scale=0.56]{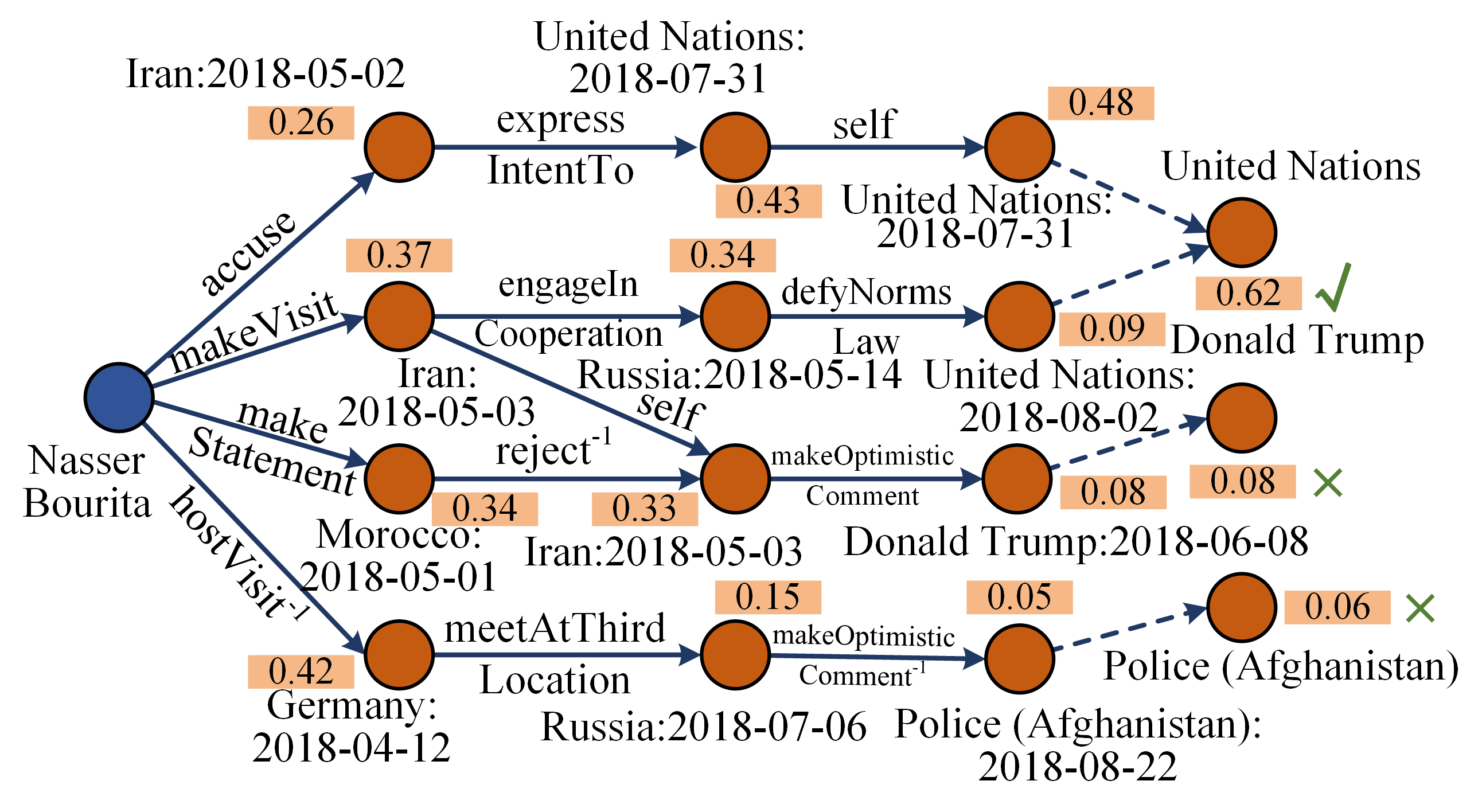}
\end{minipage}
&
\begin{minipage}{0.6\textwidth}
\begin{enumerate}[label={[\arabic{enumi}]},leftmargin=*, itemsep=1pt]
\item 0.14 \emph{accuse}\emph{(X,Y)}$:t_1$$\land$\emph{expressIntentTo}\emph{(Y,Z)}$:t_2$ $\rightarrow$\emph{\textcolor{red}{makeVisit}(X,Z)}$:t$
\item 0.09 \emph{\textcolor{red}{makeVisit}}\emph{(X,Y$_1$)}$:t_1$$\land$\emph{engageInCooperation}\emph{(Y$_1$,Y$_2$)}$:t_2$$\land$\emph{defyNormsLaw}\emph{(Y$_2$,Z)}$:t_3$\\$\rightarrow$ \emph{\textcolor{red}{makeVisit}(X,Z)}$:t$
\item 0.11 \emph{makeStatement}\emph{(X,Y$_1$)}$:t_1$$\land$\emph{reject}$^{-1}$\emph{(Y$_1$,Y$_2$)}$:t_2$$\land$\\\emph{makeOptimisticComment}\emph{(Y$_2$,Z)}$:t_3$$\rightarrow$\emph{\textcolor{red}{makeVisit}(X,Z)}$:t$
\item 0.25 \emph{\textcolor{red}{makeVisit}}\emph{(X,Y)}$:t_1$$\land$\emph{makeOptimisticComment}\emph{(Y,Z)}$:t_2$ $\rightarrow$\emph{\textcolor{red}{makeVisit}(X,Z)}$:t$
\item 0.17 \emph{hostVisit}$^{-1}$\emph{(X,Y$_1$)}$:t_1$$\land$\emph{meetAtThirdLocation}\emph{(Y$_1$,Y$_2$)}$:t_2$$\land$\\\emph{makeOptimisticComment}$^{-1}$\emph{(Y$_2$,Z)}$:t_3$$\rightarrow$ \emph{\textcolor{red}{makeVisit}(X,Z)}$:t$
\end{enumerate}
\end{minipage}\\
\multicolumn{2}{c}{\cellcolor{gray!30}Extrapolation reasoning: query (\emph{Nasser Bourita}, \emph{makeVisit}, ?, \emph{2018-09-28}) in ICEWS18}\\
\bottomrule
\end{tabular}
\label{case1}
\end{table*}

\begin{table*}[t!]
\centering
\caption{Some reasoning cases in inductive scenarios, where learned FOL rules are displayed.
Relations marked with red represent the target relation to be predicted. ``${-1}$'' denotes the reverse of a specific relation and textual descriptions of some relations are simplified.}
\begin{tabular}{c}
\toprule
\begin{minipage}{0.99\textwidth}
{[1]} 0.41 \emph{\textcolor{red}{memberMeronym}}\emph{(X,Y$_1$)}$\land$\emph{hasPart}\emph{(Y$_1$,Y$_2$)}$\land$\emph{hasPart}$^{-1}$\emph{(Y$_2$,Z)} $\rightarrow$ \emph{\textcolor{red}{memberMeronym}(X,Z)}\\
{[2]} 0.19 \emph{hasPart}$^{-1}$\emph{(X,Y$_1$)}$\land$\emph{hypernym}\emph{(Y$_1$,Y$_2$)}$\land$\emph{memberOfDomainUsage}$^{-1}$\emph{(Y$_2$,Z)} $\rightarrow$ \emph{\textcolor{red}{memberMeronym}(X,Z)}\\
{[3]} 0.25 \emph{hypernym}\emph{(X,Y$_1$)}$\land$\emph{hypernym}$^{-1}$\emph{(Y$_1$,Y$_2$)}$\land$\emph{\textcolor{red}{memberMeronym}}\emph{(Y$_2$,Z)} $\rightarrow$ \emph{\textcolor{red}{memberMeronym}(X,Z)}\\
{[4]} 0.17 \emph{hypernym}\emph{(X,Y$_1$)}$\land$\emph{hypernym}$^{-1}$\emph{(Y$_1$,Y$_2$)}$\land$\emph{hasPart}\emph{(Y$_2$,Z)} $\rightarrow$ \emph{\textcolor{red}{memberMeronym}(X,Z)}\\
\end{minipage}\\
\cellcolor{gray!30}Inductive reasoning: query (\emph{08174398}, \emph{memberMeronym}, ?) in WN18RR v3\\
\midrule
\begin{minipage}{0.99\textwidth}
{[1]} 0.32 \emph{\textcolor{red}{filmReleaseRegion}}\emph{(X,Y$_1$)}$\land$\emph{\textcolor{red}{filmReleaseRegion}}$^{-1}$\emph{(Y$_1$,Y$_2$)}$\land$\emph{filmCountry}\emph{(Y$_2$,Z)} $\rightarrow$ \emph{\textcolor{red}{filmReleaseRegion}(X,Z)}\\
{[2]} 0.10 \emph{distributorRelation}$^{-1}$\emph{(X,Y$_1$)}$\land$\emph{nominatedFor}\emph{(Y$_1$,Y$_2$)}$\land$\emph{\textcolor{red}{filmReleaseRegion}}$^{-1}$\emph{(Y$_2$,Z)} $\rightarrow$ \emph{\textcolor{red}{filmReleaseRegion}(X,Z)}\\
{[3]} 0.19 \emph{\textcolor{red}{filmReleaseRegion}}\emph{(X,Y$_1$)}$\land$\emph{exportedTo}$^{-1}$\emph{(Y$_1$,Y$_2$)}$\land$\emph{locationCountry}\emph{(Y$_2$,Z)} $\rightarrow$ \emph{\textcolor{red}{filmReleaseRegion}(X,Z)}\\
{[4]} 0.05 \emph{filmCountry}\emph{(X,Y$_1$)}$\land$\emph{\textcolor{red}{filmReleaseRegion}}$^{-1}$\emph{(Y$_1$,Y$_2$)}$\land$\emph{filmMusic}\emph{(Y$_2$,Z)} $\rightarrow$ \emph{\textcolor{red}{filmReleaseRegion}(X,Z)}\\
\end{minipage}\\
\cellcolor{gray!30}Inductive reasoning: query (\emph{/m/0j6b5}, \emph{filmReleaseRegion}, ?) in FB15k-237 v3\\
\midrule
\begin{minipage}{0.99\textwidth}
{[1]} 0.46 \emph{\textcolor{red}{collaboratesWith}}$^{-1}$\emph{(X,Z)} $\rightarrow$ \emph{\textcolor{red}{collaboratesWith}(X,Z)}\\
{[2]} 0.38 \emph{\textcolor{red}{collaboratesWith}}$^{-1}$\emph{(X,Y$_1$)}$\land$\emph{holdsOffice}\emph{(Y$_1$,Y$_2$)}$\land$\emph{holdsOffice}$^{-1}$\emph{(Y$_2$,Z)} $\rightarrow$ \emph{\textcolor{red}{collaboratesWith}(X,Z)}\\
{[3]} 0.03 \emph{\textcolor{red}{collaboratesWith}}$^{-1}$\emph{(X,Y$_1$)}$\land$\emph{graduatedFrom}\emph{(Y$_1$,Y$_2$)}$\land$\emph{graduatedFrom}$^{-1}$\emph{(Y$_2$,Z)} $\rightarrow$ \emph{\textcolor{red}{collaboratesWith}(X,Z)}\\
{[4]} 0.03 \emph{\textcolor{red}{collaboratesWith}}$^{-1}$\emph{(X,Y$_1$)}$\land$\emph{\textcolor{red}{collaboratesWith}}\emph{(Y$_1$,Y$_2$)}$\land$\emph{graduatedFrom}\emph{(Y$_2$,Z)} $\rightarrow$ \emph{\textcolor{red}{collaboratesWith}(X,Z)}\\
\end{minipage}\\
\cellcolor{gray!30}Inductive reasoning: query (\emph{Hillary Clinton}, \emph{collaboratesWith}, ?) in NELL v3\\
\bottomrule
\end{tabular}
\label{case2}
\end{table*}

\subsection{Case Studies (RQ4)}

To show the actual reasoning process of \textsc{Tunsr}, some practical cases are presented in detail on all four reasoning scenarios,
which illustrate the transparency and interpretability of the proposed \textsc{Tunsr}.
For better presentation, the maximum length of the reasoning iterations is set to 3.
Specifically, Table~\ref{case1} shows the reasoning graphs for three specific queries on transductive, interpolation, and extrapolation scenarios, respectively,
The propositional attention weights of nodes are listed near them, which represent the propositional reasoning score of each node at the current step.
For example, in the first case, the uppermost propositional reasoning path (\emph{00238867}, \emph{verbGroup}$^{-1}$, \emph{00239321}) at first step learns a large attention score for the correct answer \emph{00239321}. Generally, nodes with more preceding neighbors or larger preceding attention weights significantly impact subsequent steps and the prediction of final entity scores.
Besides, we observe that propositional and first-order reasoning have an incompletely consistent effect.
For example, the FOL rules of ``[3]'' and ``[4]'' in the third case have relatively high rule confidence values compared with ``[1]'' and ``[2]'' (0.11, 0.25 vs. 0.14, 0.09),
but the combination of their corresponding propositional reasoning paths ``(\emph{Nasser Bourita}, \emph{makeStatement}, \emph{Morocco:2018-05-01}, \emph{reject}$^{-1}$, \emph{Iran:2018-05-03}, \emph{makeOptimisticComment}, \emph{Donald Trump:2018-06-08})'' and ``(\emph{Nasser Bourita}, \emph{makeVisit}, \emph{Iran:2018-05-03}, \emph{self}, \emph{Iran:2018-05-03}, \emph{makeOptimisticComment}, \emph{Donald Trump:2018-06-08})'' has a small propositional attention, i.e., 0.08.
This combination prevents the model from predicting the wrong answer \emph{Donald Trump}.
Thus, propositional and FOL reasoning can be integrated to jointly guide the reasoning process, leading to more accurate reasoning results.

Table~\ref{case2} shows some learned FOL rules of inductive reasoning on WN18RR v3, FB15k-237 v3, and NELL v3 datasets.
As the inductive setting is entity-independent, so the propositional reasoning part is not involved here.
Each rule presented carries practical significance and is readily understandable for humans.
For instance, rule ``[1]'' \emph{collaboratesWith}$^{-1}$\emph{(X, Z)} $\rightarrow$ \emph{collaboratesWith(X, Z)} in the third case has a relatively high confidence value (0.46).
This aligns with human commonsense cognition, as the relation \emph{collaboratesWith} has mutual characteristics for subject and object and can be derived from each other.
If \emph{person a} has collaborated with \emph{person b}, it inherently implies \emph{person b} has collaborated with \emph{person a}.
These results illustrate the effectiveness of the rules learned by \textsc{Tunsr} and its interpretable reasoning process.

\section{Conclusion and Future Works}

To combine the advantages of connectionism and symbolicism of AI for KG reasoning, we propose a unified neurosymbolic framework \textsc{Tunsr} for both perspectives of methodology and reasoning scenarios, including transductive,
inductive, interpolation, and extrapolation reasoning. 
\textsc{Tunsr} first introduces a consistent structure of reasoning graph that starts from the query entity and constantly expands subsequent nodes by iteratively searching posterior neighbors.
Based on it, a forward logical message-passing mechanism is proposed to update both the propositional representations and attentions, as well as FOL representations and attentions of each node in the expanding reasoning graph.
In this way, \textsc{Tunsr} conducts the transformation of merging multiple rules by merging possible relations at each step by using FOL attentions.
Through gradually adding rule bodies and updating rule confidence, the real FOL rules can be easily induced to constantly perform attention calculation over the reasoning graph, which is summarized as the FARI algorithm.
The experiments on 19 datasets of four different reasoning scenarios illustrate the effectiveness of \textsc{Tunsr}.
Meanwhile, the ablation studies show that propositional and FOL have different impacts.
Thus, they can be integrated to improve the whole reasoning results.
The case studies also verify the transparency and interpretability of its computation process.

The future works lie in two folds.
Firstly, we aim to extend the application of this idea to various reasoning domains, particularly those necessitating interpretability for decision-making~\cite{cambria2023seven}, such as intelligent healthcare and finance.
We anticipate this will enhance the accuracy of reasoning while simultaneously offering human-understandable logical rules as evidence.
Secondly, we intend to integrate the concept of unified reasoning with state-of-the-art technologies to achieve optimal results.
For instance, large language models have achieved great success in the community of natural language processing and AI, while they often encounter challenges when confronted with complex reasoning tasks~\cite{xu2023large}.
Hence, there is considerable prospect for large language models to enhance reasoning capabilities.



{\small
\bibliographystyle{revision_ref}
\bibliography{main}
}

\ifCLASSOPTIONcaptionsoff
  \newpage
\fi

\vfill


\end{document}